\documentclass[lettersize,journal]{IEEEtran}

\usepackage[export]{adjustbox}
\usepackage{amsmath,amsfonts,amssymb,mathtools,amsthm}
\usepackage{array}
\usepackage[caption=false,font=normalsize,labelfont=sf,textfont=sf]{subfig}
\usepackage{textcomp}
\usepackage{stfloats}
\usepackage{url}
\usepackage{verbatim}

\usepackage{graphicx}
\usepackage{booktabs} 
\usepackage{hyperref}

\theoremstyle{plain}

\theoremstyle{definition}

\theoremstyle{remark}

\newcommand{\hide}[1]{}
\newcommand{\G}{\mathcal{G}}
\usepackage{overpic}
\usepackage{multirow}
\usepackage{booktabs}
\usepackage[inline]{enumitem}
\newcolumntype{C}{>{\centering\arraybackslash}m{2cm}}

\def\BibTeX{{\rm B\kern-.05em{\sc i\kern-.025em b}\kern-.08em T\kern-.1667em\lower.7ex\hbox{E}\kern-.125emX}}
\usepackage{balance}

\usepackage{xcolor}
\newcommand{\gM}{\mathcal{M}}
\newcommand{\red}[1]{\textcolor{black}{#1}}

\newcommand{\etal}{\textit{et al}. }
\newcommand{\ie}{\textit{i}.\textit{e}., }
\newcommand{\eg}{\textit{e}.\textit{g}., }

\newcommand{\wrt}{with respect to }

\begin{document}

\title{Graph Kernel Neural Networks}

\author{Luca Cosmo, Giorgia Minello, Alessandro Bicciato, Michael M. Bronstein, Emanuele Rodolà,\\ Luca Rossi, Andrea Torsello
\thanks{L. Cosmo, G. Minello, A. Bicciato, and A. Torsello are with Ca' Foscari University of Venice, Venice, Italy. M. M. Bronstein is with Oxford University, United Kingdom. E. Rodolà is with Sapienza University of Rome, Rome, Italy. L. Rossi is with The Hong Kong Polytechnic University, Hong Kong. Corresponding author: luca.rossi@polyu.edu.hk}}

\markboth{Journal of \LaTeX\ Class Files,~Vol.~14, No.~8, August~2021}%
{Shell \MakeLowercase{\textit{et al.}}: A Sample Article Using IEEEtran.cls for IEEE Journals}

\IEEEpubid{0000--0000/00\$00.00~\copyright~2021 IEEE}

\maketitle

\pagestyle{empty}
\IEEEoverridecommandlockouts
\makeatletter
\def\ps@IEEEtitlepagestyle{%
\def\@oddhead{}%
\def\@evenhead{}%
\def\@oddfoot{}%
\def\@evenfoot{}}
\makeatother
\IEEEpubidadjcol
\IEEEpubid{%
    \begin{minipage}{\textwidth}\ \\[12pt] \end{minipage}
}

\begin{abstract}
The convolution operator at the core of many modern neural architectures can effectively be seen as performing a dot product between an input matrix and a filter. While this is readily applicable to data such as images, which can be represented as regular grids in the Euclidean space, extending the convolution operator to work on graphs proves more challenging, due to their irregular structure. In this paper, we propose to use graph kernels, \ie kernel functions that compute an inner product on graphs, to extend the standard convolution operator to the graph domain. This allows us to define an entirely structural model that does not require computing the embedding of the input graph. Our architecture allows to plug-in any type of graph kernels and has the added benefit of providing some interpretability in terms of the structural masks that are learned during the training process, similarly to what happens for convolutional masks in traditional convolutional neural networks. We perform an extensive ablation study to investigate the model hyper-parameters' impact and show that our model achieves competitive performance on standard graph classification and regression datasets.
\end{abstract}


\section{Introduction}
\IEEEPARstart{I}n recent years, graph neural networks (GNNs) have gained increasing traction in the machine learning community. Graphs have long been used as a powerful abstraction for a wide variety of real-world data where structure plays a key role~\cite{duin2012dissimilarity}, from collaborations~\cite{lima2014coding,kipf2016semi} to biological~\cite{gilmer2017neural,ye2015thermodynamic} and physical~\cite{shlomi2020graph} data, to mention a few. Before the advent of GNNs, {\em graph kernels} provided a principled way to deal with graph data in the traditional machine learning setting~\cite{shervashidze2011weisfeiler,bai2015quantum,minello2019can}. However, with the attention of machine learning researchers steadily shifting away from hand-crafted features toward end-to-end models where both the features and the model are learned together, neural networks have quickly overtaken kernels as the framework of choice to deal with graph data, leading to multiple popular architectures such as~\cite{kipf2016semi,gilmer2017neural,velickovic2018graph}.

The main obstacle that both traditional and deep learning methods have to overcome when dealing with graph data is also the source of interest for using graphs as data representations. The richness of graphs means that there is no obvious way to embed them into a vector space, a necessary step when the learning method expects vectors in input. One reason is the lack of a canonical ordering of the nodes in a graph, requiring either permutation-invariant operations or an alignment to a reference structure. Moreover, even if the order or correspondence can be established, the dimension of the embedding space may vary due to structural modifications, \ie changes in the number of nodes and edges.

In traditional machine learning, kernel methods, and particularly graph kernels, provide an elegant way to sidestep this issue by replacing explicit vector representations of the data points with a positive semi-definite matrix of their inner products. Thus, any algorithm that can be formulated in terms of scalar products between input vectors can be applied to a set of data (such as graphs) on which a kernel is defined.

GNNs apply a form of generalized convolution operation to graphs, which can be seen as a message-passing strategy where the node features are propagated over the graph to capture node interactions. Compared to standard convolutional neural networks (CNNs), the main drawback of these architectures is the lack of a straightforward way to interpret the dependency between the output and the presence of certain features and structural patterns in the input graph.
In this paper, we propose a neural architecture that bridges the two worlds of graph kernels and GNNs. The key intuition underpinning our work is that there exists an analogy between the traditional convolution operator, which can be seen as performing an inner product between an input matrix and a filter, and graph kernels, which compute an inner product of graphs. 
As such, graph kernels provide the natural tool to generalize the concept of convolution to the graph domain, which we term \emph{graph kernel convolution} (GKC). Given a graph kernel of choice, in each GKC layer, the input graph is compared against a series of structural masks (analogous to the convolutional masks in CNNs), which effectively represent learnable subgraphs. These, in turn, can offer better interpretability in the form of insights into what structural patterns in the input graphs are related to the corresponding output (\eg the carcinogenicity of a chemical compound).
\IEEEpubidadjcol

\textbf{Our main contributions are:}
\begin{itemize}
\item We introduce a neural model that is fully structural, unlike existing approaches that require embedding the input graph into a larger, relaxed space. While the latter allows one to seek more complex and arbitrary decision boundaries, it also has the potential of overfitting the problem and creating more local optima, as evidenced by the need to use dropout strategies to avoid overfitting in Graph Convolutional Networks (GCNs)~\cite{rong2019dropedge,cong2021provable}.
\item In order to cope with the non-differentiable nature of graphs, we let the learned structural masks encode distributions over graphs, which in turn allows us to compute the gradient of the expected kernel response between the learned masks and the input subgraphs. This is, in turn, achieved through a computationally efficient importance sampling approach.
\item Our architecture allows to plug-in any type of graph kernel (not just differentiable kernels as in~\cite{nikolentzos2020random,feng2022kergnns}).
\item As an added benefit, our model provides some interpretability regarding the structural masks learned during the training process, similarly to convolutional masks in traditional CNNs.
\item Finally, we analyze the expressive power of our model, and we argue that it is greater than that of standard message-passing GNNs and the equivalent Weisfeiler–Lehman (WL) graph isomorphism test.
\end{itemize}

The fundamental goal of our network is to reconstruct the structural information needed for classification and is thus particularly suited for problems where the structure plays a pivotal role. These are often graph classification and regression problems, where structure provides the most relevant information for classification, while node/edge features take a secondary role. The situation is usually reversed for node classification problems, where the distribution of features across the immediate neighbors of a node holds most of the information needed to classify the node itself. For these reasons, in this paper, we focus on graph classification and regression problems. However, we stress that our model can indeed be extended to tackle node-level tasks.

The remainder of this paper is organized as follows. In Section~\ref{sec:related}, we review the related work. In Section~\ref{sec:our}, we present our model, where graph kernels are used to redefine convolution on graph data. In Section~\ref{sec:experiments}, we investigate the hyper-parameters of our architecture through an extensive ablation study, provide some insight into the interpretability of the structural masks, and evaluate our architecture on standard graph classification and regression benchmarks. Finally, Section~\ref{sec:conclusion} concludes the paper.
\section{Related work}\label{sec:related}
The majority of graph kernels belong to one of two main categories: 1) bag-of-structures and 2) information propagation kernels. Bag-of-structures kernels compute the similarity between a pair of input graphs by first decomposing them into simpler substructures and then counting the number of isomorphic substructures between the two input graphs. Depending on the type of substructure considered, one can build a multitude of different kernels, \eg subtrees~\cite{ramon2003expressivity}, shortest paths~\cite{borgwardt2005shortest}, and graphlets~\cite{shervashidze2009efficient}. Information propagation kernels, on the other hand, include methods where pairs of input graphs are compared based on how information diffuses on them. Examples include random walk kernels~\cite{kashima2003marginalized,bai2013graph}, quantum walk kernels~\cite{bai2015quantum,rossi2015measuring,zhang2020r}, and kernels based on iterative label refinements~\cite{shervashidze2011weisfeiler}. While some kernels work only on undirected and unattributed graphs, other kernels are designed to handle attributes as well, either discrete- or continuous-valued~\cite{shervashidze2011weisfeiler,da2017tree}. For a detailed review and historical perspective on graph kernels, we refer the reader to the recent survey of Kriege \etal~\cite{kriege2020survey}.

In recent years, with the advent of deep learning and the renewed interest in neural architectures, the focus of graph-based machine learning researchers has quickly moved to extending deep learning approaches to deal with graph data. Fundamentally, the principle underpinning most GNNs is that of exploiting the structure of the graph to propagate the node feature information iteratively. One of the first papers to propose the idea of GNNs is that of Scarselli \etal~\cite{scarselli2008graph}, where an information diffusion mechanism is used to learn the nodes' latent representations by exchanging neighborhood information. Sperduti and Starita~\cite{sperduti1997supervised} and Micheli~\cite{micheli2009neural} had previously relied instead on recursive operators, which can be seen as a variation of convolutional operators where different layers share the same weights, to adapt neural networks to operate on structured data such as graphs. Most modern GNNs, however, appear to fall within three main categories, as discussed by Bronstein \etal~\cite{bronstein2021geometric}: 1) convolutional~\cite{kipf2016semi,atwood2016diffusion,levie2018cayleynets,bai2020learning,li2021learning}, 2) attentional~\cite{velickovic2018graph}, and 3) message passing~\cite{gilmer2017neural}. The message-passing model is often seen as the most general one and has been shown to be formally equivalent to the WL graph isomorphism test under some technical conditions~\cite{xu2018powerful,morris2019weisfeiler}.

An increasing number of researchers are trying to overcome the latter limitation by moving away from standard message-passing GNNs~\cite{bicciato2022classifying,bicciato2024gnn}. Eliasof \etal~\cite{eliasof2021pde} propose a reinterpretation of graph convolution in terms of partial differential equations on graphs, noting that different problems can benefit from different networks dynamics, hence suggesting the need to look beyond diffusion. Bevilaqua \etal~\cite{bevilacqua2022equivariant} represent graphs as bags of substructures and show that this allows the WL test to distinguish between otherwise indistinguishable graphs. Bouritsas \etal~\cite{bouritsas2022improving} take instead an alternative approach where augmenting the graph nodes attributes with positional encoding is shown to improve the ability of the WL test to discriminate between non-isomorphic graphs. The field of GNNs is, in general, in rapid and continuous expansion, so presenting a comprehensive review of the literature is beyond the scope of this paper. Instead, we refer the interested readers to~\cite{wu2020comprehensive,bronstein2021geometric}.

In this work, we argue that a natural extension of the convolution operation to the graph domain, and thus an ideal candidate to build graph CNNs, already exists in the form of graph kernels. A number of works in the literature have investigated potential synergies between GNNs and kernels. Lei \etal~\cite{lei2017deriving} introduce a class of deep recurrent neural architectures to show that this lies in the reproducing kernel Hilbert space (RKHS) of graph kernels. Nikolentzos \etal~\cite{nikolentzos2018kernel} compute continuous embeddings of graphs using kernels and plug them into a neural network. Xu \etal~\cite{xu2018powerful} and Morris \etal~\cite{morris2019weisfeiler} show that GNNs have the same expressiveness as the WL graph kernel~\cite{shervashidze2011weisfeiler} and propose a new generalized architecture with increased expressive power. Du \etal~\cite{du2019graph} take a somewhat different approach and exploit neural networks to introduce a new graph kernel. This, in turn, is shown to be equivalent to an infinitely wide GNN initialized with random weights and trained with gradient descent. Chen \etal~\cite{chen2020convolutional} propose a graph neural architecture where each layer enumerates local substructures around each node and then maps them to an RKHS via a Gaussian kernel mapping. In the context of graph compression, Bouritsas \etal~\cite{bouritsas2021partition} learn how to best decompose the graph into small substructures that are also learned. Navarin \etal~\cite{navarin2020learning} propose a modification of DGCNN~\cite{zhang2018end}, a popular GNN architecture, where a multi-task learning approach is used to drive the learned node embeddings to be close to those computed by graph kernels.
\section{GNNs from Graph Kernels}\label{sec:our}
In this section we introduce the proposed neural architecture, the Graph Kernel Neural Network (GKNN). Before that, however, we introduce some background concepts that will help better understand the proposed model.
\subsection{Preliminaries}
\textit{\textbf{Kernel methods.}}
This class of algorithms is capable of learning when presented with a particular positive pairwise measure on the input data, known as a kernel. Consider a set $X$ and a positive semi-definite kernel function $\mathcal{K}: X \times X \rightarrow \mathbb{R}$ such that there exists a map $\phi: X \rightarrow H$ into a Hilbert space $H$ and $\mathcal{K}(x,y) = \phi(x)^\top \phi(y)$ for all $x,y \in X$. Crucially, $X$ can represent any set of data on which a kernel can be defined, from $\mathbb{R}^d$ to a finite set of graphs. Hence, the field of machine learning is ripe with examples of graph kernels (see Section~\ref{sec:related}), which are nothing but positive semi-definite pairwise similarity measures\footnote{Note that while several kernels can be interpreted as similarity measures, this is not always the case~\cite{nebel2017types}.} on graphs. These can be either implicit (only $\mathcal{K}$ is computed) or explicit ($\phi$ is also computed).

\textit{\textbf{Weisfeiler-Lehman test.}}
The WL kernel~\cite{shervashidze2011weisfeiler} is one of the most powerful and commonly used graph kernels, and it is based on the 1-dimensional Weisfeiler-Lehman (1-WL) graph isomorphism test. The idea underpinning the test is to partition the node set by iteratively propagating the node labels between adjacent nodes. With each iteration, the set of labels accumulated at each node of the graph is then mapped to a new label through a hash function. This procedure is repeated until the cardinality of the set of labels stops growing. Two graphs can then be compared in terms of their label sets at convergence, with two graphs being isomorphic only if their label sets coincide. 

\textit{\textbf{Learning on graphs.}}
\red{Let $\G=(V,X,E)$ be an undirected graph with $|V|$ nodes and $|E|$ edges, where each node $v$ is associated to a label $x(v)$ belonging to a dictionary $\mathbf{D}$. Note that, generally, graphs can be directed and have continuous and discrete attributes on both the nodes and edges; however, for simplicity in this paper, we restrict our attention to undirected node-labeled graphs. Indeed, the model described in this paper can be easily applied to directed, edge-labeled graphs as well. A common goal in graph machine learning problems is to produce a vector representation of $\G$ that is aware of both the node labels and the structural information of $\G$. Generalizing the convolution operator to graphs, GNNs learn the parameters $\Theta$ of a function $h$ that performs message passing \cite{gilmer2017neural} in the 1-hop neighborhood of each node $v \in V$,}
\begin{equation}
z(v) =\sum_{u \in \mathcal{N}_{\G}^1 (v)} h_{\Theta}\left(x(v), x(u) \right) \\,
\end{equation}
where $\mathcal{N}_{\G}^1 (v)$ denotes the 1-hop neighborhood of $v$.
The convolution is usually repeated for $L$ layers, and the final vectorial representation \red{$Z_{pool}$} is obtained by applying a node-wise permutation invariant aggregation operator $\square$ on the node features $z(v)$.

\subsection{Proposed architecture}
\begin{figure}[t!]
\centering
\begin{overpic}
[trim=0cm 0cm 0cm 0cm,clip,width=1\linewidth]{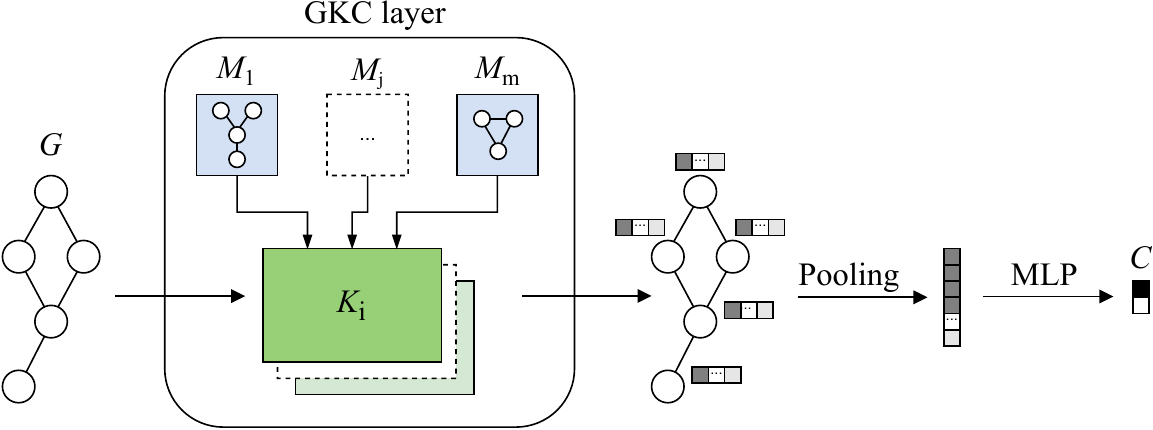}
\end{overpic}
\vskip -1mm
\caption{The proposed GKNN architecture. The input graph is fed into one or more GKC layers, where subgraphs centered at each node are compared to a series of structural masks through a kernel function. The output is a new set of real-valued feature vectors associated to the graph nodes. We obtain a graph-level feature vector through pooling on the nodes features, which is then fed to an MLP to output the final classification label.}
\label{fig:model}
\end{figure}

The core feature of the proposed model is the definition of a new approach to perform the convolution operation on graphs. Unlike the diffusion process performed by message passing techniques, we design the {\em Graph Kernel Convolution} (GKC) operation in terms of the inner product between graphs computed through a graph kernel function.

Given a graph $\G=(V,X,E)$ with $|V|$ nodes, we extract $|V|$ subgraphs $\mathcal{N_G}^r(v)$ of radius $r$, centered at each vertex $v \in V$. Each subgraph consists of the central node $v$, the nodes at distance at most $r$ from $v$, and the edges between them. Each such subgraph is then compared to a set of \red{learnable} structural masks $\{ \mathcal{M}_1, \hdots, \mathcal{M}_m\}$ through
\begin{equation}
z_i(v) = \mathcal{K}( \mathcal{M}_i, \mathcal{N}_{\mathcal{G}}^r(v)),
\label{eq:convolution}
\end{equation}
where the resulting feature vector $z(v)$ is a non-negative real-valued $m$-dimensional vector collecting the kernel responses, and the $i$-th structural mask 
\red{$\mathcal{M}_i= (V_{\mathcal{M}_i},X_{\mathcal{M}_i},E_{\mathcal{M}_i})$ }
is a first-order stochastic graph, \ie a distribution over graphs with node set 
\red{$V_{\mathcal{M}_i}$} where the edges $e\in E_{\mathcal{M}_i}$ are sampled from independent Bernoulli trials each with its own parameter $\theta_e$ \red{and each node label $x\in X_{\mathcal{M}_i}$ is sampled from the corresponding discrete probability distribution over a dictionary $\mathbf{D}$ with parameters $\phi_{xd} \forall d \in \mathbf{D}$ each expressing the probability of $x$ assuming value $d$.
} \red{The probability distributions $\theta$ and $\phi$ over edges and node labels defining each mask are effectively the learnable parameters of the GKC layer.}

\textit{\textbf{Multi-layer architecture.}}
The output node feature $z(v)\in\mathbb{R}_+^m$ of a GKC operation consists of a continuous vector containing the graph kernel responses between the structural masks and the subgraph centered on the node $v$. For graph kernels functions requiring discrete labels, the output feature vector thus needs to be discretized before giving it as input to the next layer. In this case, we interpret $z(v)$ as an unnormalized probability distribution over a set of $m$ labels from which we sample the new label $x(v)$ for the node $v$ at the next GKC layer.

Since in this paper we focus on graph-level tasks, the final graph embedding after $L$ layers is obtained by $\square(Z^1| \dots |Z^L)$, where $Z^\ell$ are the output features of the $\ell$th layer and $\cdot|\cdot$ indicates the concatenation of node-wise features. This is then fed to a multilayer perceptron (MLP) layer to obtain the final classification. In particular, in all our experiments we use sum pooling as the aggregation operator $\square$. Figure~\ref{fig:model} shows an overview of the proposed multilayer architecture.

\red{\textit{\textbf{Cross-entropy loss.}}
Our learning problem can be formulated as follows: given a set $\{\G_1,\dots,\G_B\}$ of $B$ training input graphs with node labels belonging to the dictionary $\mathbf{D}$ and associated class labels $y_1 \dots y_B$, our goal is to find the optimal parameters $\Phi$ and the masks $\mathcal{M}_i$ with corresponding edge observation probabilities $\Theta$ of the model $g$ that minimize the cross-entropy loss
\begin{equation}\label{eq:cross_entropy}
loss_{CE} = \sum_{i=1}^B \text{\textit{cross-entropy}}(\mathbb{E}_{\Theta}[g_{\mathcal{M}_1 \dots \mathcal{M}_m, \Phi}(\G_i)], y_i )\,.
\end{equation}}

\textit{\textbf{Jensen–Shannon divergence loss.}}
Depending on the number of structural masks to learn, we experimentally observed that different structural masks can give a similar response over the same node. To mitigate this behavior and push the model to learn a more descriptive node feature vector, we propose to regularize the structural masks learning process by adding a Jensen–Shannon divergence (JSD) loss. The JSD is computed between the feature dimensions considered as probability distributions over the nodes of the graph. 

Let $ P_i = \{\alpha z_i(v) | v \in V\}$ be the probability distribution induced by the $i$-th mask over the graph nodes. $\alpha$ is a scaling factor ensuring that $\sum_v P_i(v) = 1$. We define the JSD loss as
\begin{equation}
loss_{JSD} = -H\left(\sum_{i=1}^{m} P_{i}\right)+\sum_{i=1}^{m} H\left(P_{i}\right) \,,
\end{equation}
where $H(P)$ is the Shannon entropy. The final loss we optimize is the sum of 1) the cross-entropy loss of Eq.~\ref{eq:cross_entropy} 2) $loss_{JSD}$ multiplied by a weighting factor \red{$\lambda$, i.e.,
\begin{equation}
loss = loss_{CE}+\lambda loss_{JSD} \,.
\end{equation}
Note that while previous work in the literature used the JSD to generalize the widely used cross-entropy loss function~\cite{englesson2021generalized}, here we use it as a regularization term. That is, rather than minimizing the JSD or cross-entropy in order to limit the distance to a target class~\cite{bai2019quantum}, we maximize the JSD to force the learned distributions to be as far away from one another as possible.}

\subsection{Optimization strategy}
When optimizing the structural masks, we need to consider two main challenges. First, the number of nodes of the subgraphs is not fixed and can, in principle, vary from $1$
to $n$.
Second, since graph kernel functions are not in general differentiable, the automatic differentiation mechanism of common neural network optimization libraries cannot be directly applied to our model. 

\textit{\textbf{Structural masks representation.}}
When defining the space of graphs on which we want to optimize the structural masks $\mathcal{M}$, we have to consider the possible substructures present in the input graph that characterize it as belonging to a specific class. Assuming that this knowledge is not known {\em a priori}, we should allow to learn structures as large as the graph itself. Unfortunately, since the space of graphs grows exponentially with the number of nodes, this would be impractical. Moreover, under the assumption of the presence of localized characterizing substructures, we usually need graphs of few nodes to capture their presence. In our implementation, we fix a maximum number of nodes $p_{max}$ for each substructure and optimize for structural masks in the space
\begin{equation}\label{eq:min}
\mathcal{M} \in \bigcup_{p=1}^{p_{max}} \left ( \mathbf{\tilde{G}}_p \times \mathbf{D}^p \right )\,,
\end{equation}
where $\mathbf{\tilde{G}}_p$ indicates the set of all possible connected graphs of $p$ nodes, and $\mathbf{D}^p$ the labeling space of $p$ nodes. The impact of this hyper-parameter is studied in the ablation study in Section~\ref{sec:experiments}. 
\red{With a slight abuse in notation, here $\mathcal{M}$ represents a discrete graph with discrete node labels, but we will relax it to its stochastic version in the subsequent paragraph.}

\textit{\textbf{Structural masks optimization.}}
Graph kernels, as functions operating on discrete graph structures, are in general not differentiable. In order to be able to optimize the structural masks, we relax them to be distributions over graphs and take the expectation of the kernel over said distributions. Namely, let 
\red{$\mathcal{M}_i= (V_{\mathcal{M}_i},X_{\mathcal{M}_i},E_{\mathcal{M}_i})$ }
be our {\em i}-th structural mask, where \red{$V_{\mathcal{M}_i}$ is the set of $p$ nodes, $X_{\mathcal{M}_i}$ is the set of $p$ discrete probability distributions over the $|D|$ labels,} and $E_{\mathcal{M}_i}$ is the edge observation models consisting of $\binom{n}{2}$ independent Bernoulli trials with (learnable) parameters $\theta_1,\ldots,\theta_{\binom{n}{2}}$.

\red{A sample $G\sim \mathcal{M}_i $  is a graph $G=(V,X,E)$} whose edge set $E$ is the result of sampling from the Bernoulli trials to decide whether the corresponding edge is present, \red{and labels $X$ are individually sampled from the corresponding node label distributions}. With this model to hand, we take as the kernel response between the subgraph $\mathcal{N}_{\mathcal{G}}^r(v)$ and the mask $\mathcal{M}_i$ the expectation of the kernel over the distribution of graphs associated with $\mathcal{M}_i$, \ie

\red{
\begin{equation}
z_i(v)=\mathbb{E}_{G\sim\mathcal{M}_i}[\mathcal{K}(G,\mathcal{N}_{\mathcal{G}}^r(v))]\,.
\end{equation}}

The gradient is then taken \wrt the edge observation parameters $\Theta$.

Clearly, computing the full expectation for the kernel response and its gradient is computationally too demanding. For this reason, in the next section we discuss a sampling strategy for both.

\subsection{Expected kernel estimation through importance sampling}\label{sec:importance}
In order to efficiently compute the kernel responses and their gradient \wrt the edge observation parameters $\Theta$ we adopt an importance sampling strategy. Importance sampling is a Monte Carlo method for evaluating the expectation of a function over a particular distribution while only having samples generated from a different distribution than the distribution of interest. More formally, assume you have a function $f:X->\mathbb{R}$ over a domain $X$ and two distributions $p$ and $g$ over such domain. Then we have
\begin{multline}\label{eq:importance_sampling}
\mathbb{E}_p[f] = \int_X f(x)\,p(x)dx = \int_X f(x)\frac{p(x)}{g(x)}\,g(x)dx = \\ \mathbb{E}_g\left[f(x)\frac{p(x)}{g(x)}\right]\,.
\end{multline}

Eq.~\ref{eq:importance_sampling} suggests that the expectation of $f$ \wrt the distribution $p$ can be obtained from the expectation \wrt the distribution $g$ by adding the correction term $\frac{f(x)}{g(x)}$. Importance sampling is often used to reduce the sampling variance. Indeed, if $g(x) = \frac{1}{\mathbb{E}_p[f]} f(x)p(x)$, then a single sample is sufficient to estimate $\mathbb{E}_p[f]$.
More in general, if the samples are concentrated where both the response $f$ and the density $p$ are high, then the sampling variance is greatly reduced. To this end, and observing that our first-order random graph model associated with each mask $\mathcal{M}_i$ is strongly unimodal being the combination of independent Bernoulli trials, we concentrate our sampling around the mode of the distribution. In particular, we sample from the distribution of graphs associated with the mask $\mathcal{M}_i$ the mode $G_M$ and the graphs that can be obtained from the mode with a single edge edit operation, \ie the neighboring graphs. To this end, recall that in our setting the mode is the graph such that \red{edge} $e$ is present if and only if $\theta_e>0.5$. Letting $p_e = \max(\theta_e,1-\theta_e)$, we have that $p_e$ is the probability that edge $e$ is sampled as present or absent as in the mode $G_M$ and thus the probability of sampling $G_M$ is
\begin{equation}
P(G_M) = \prod_e p_e
\end{equation}
and the density of any neighbouring graph $G_e$ obtained by editing the edge $e$ is
\begin{equation}
P(G_e) = P(G_M) \frac{1-p_e}{p_e}\,.
\end{equation}
Assuming that we are sampling uniformly from the set consisting of the mode and its neighboring graphs, our sampling density is $\frac{1}{\binom{n}{2} + 1}$ over that set, 0 otherwise.

\begin{figure}[t!]
\centering
\begin{overpic} [trim=-0cm -0cm -0cm -0.7cm, clip,  width=1\linewidth]{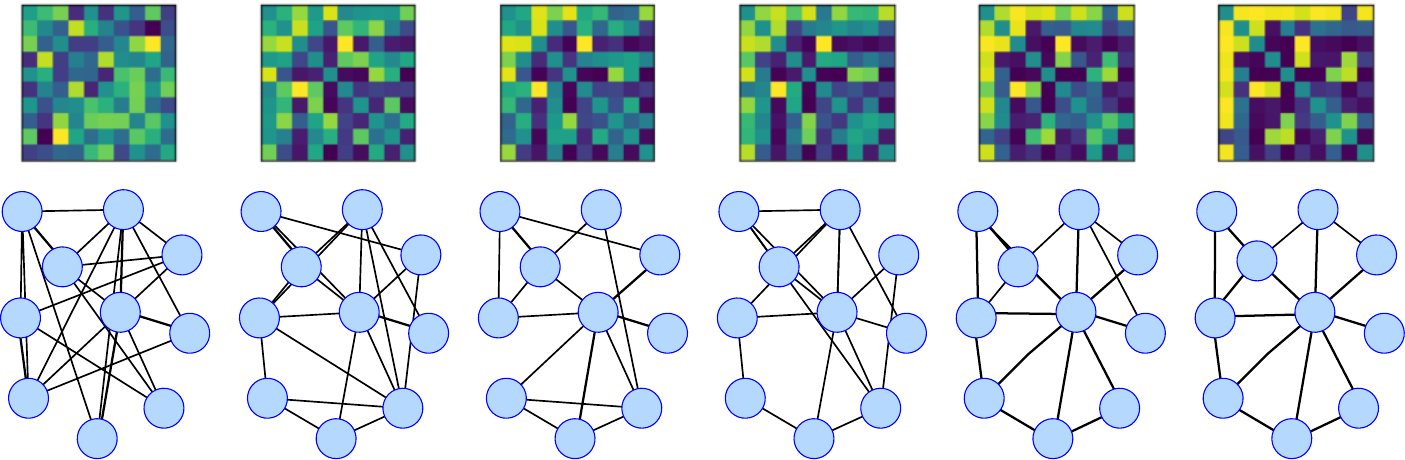}
\put(4,33){\tiny Init}
\put(20,33){\tiny Epoch 100}
\put(37,33){\tiny Epoch 200}
\put(54,33){\tiny Epoch 300}
\put(71,33){\tiny Epoch 400}
\put(90,33){\tiny Final}
\end{overpic}
\caption{Example of model training. Top: the matrix of edge observation probabilities associated to the mask $\mathcal{M}_i$ for successive epochs. Bottom: the corresponding mode graphs.}  
\label{fig:prob_evolution}
\end{figure}

With this setup, we can estimate the expected kernel response as
\red{
\begin{multline}
z_i(v)=\mathbb{E}_{G\sim\mathcal{M}_i}[\mathcal{K}(G,\mathcal{N}_{\mathcal{G}}^r(v))]\approx \\
\frac{\binom{n}{2} + 1 }{|\lbrace G \rbrace |}\sum_{G\sim\mathcal{M}_i} P(G)\mathcal{K}(G,\mathcal{N}_{\mathcal{G}}^r(v)) = \\ 
\frac{P(G_M)\left( \binom{n}{2} + 1 \right)}{|\lbrace G \rbrace |} \Bigg( \mathcal{K}(G_M,\mathcal{N}_{\mathcal{G}}^r(v)) \\
+ \sum_e \mathcal{K}(G_e,\mathcal{N}_{\mathcal{G}}^r(v))\frac{1-p_e}{p_e}\Bigg)
\end{multline}}
where $\lbrace G \rbrace$ is the set of samples, $P(G)$ is the sampling probability, $G_M$ is the mode graph, and the summation over $e$ refers to the summation over the sampled neighbouring graphs $G_e$, where $e$ is the edge that differs from the mode. Note that the first term is due to the mode. If the mode is not among the samples that term can be dropped.

Similarly, we differentiate the expectation as follows:
\begin{multline}
\frac{\partial z_i(v)}{\partial\theta_{e'}} \approx 
\frac{\left( \binom{n}{2} + 1 \right)}{|\lbrace G \rbrace |} \Bigg(  \mathcal{K}(G_M,\mathcal{N}_{\mathcal{G}}^r(v))\frac{P(G_M)}{p_{e'}}\frac{\partial p_{e'}}{\partial \theta_{e'}} +  \\ 
\sum_{e \neq e'} \mathcal{K}(G_e,\mathcal{N}_{\mathcal{G}}^r(v))\frac{P(G_e)}{p_{e'}}\frac{\partial p_{e'}}{\partial \theta_{e'}}\\
+  \mathcal{K}(G_{e'},\mathcal{N}_{\mathcal{G}}^r(v))\frac{P(x_{e'})}{1-p_{e'}}\frac{\partial (1-p_{e'})}{\partial \theta_{e'}}  \Bigg)\,.
\end{multline}
Here too, we have a term associated with the mode, plus one associated with the neighbor obtained by editing the edge $e'$. Either term can be dropped if the corresponding graph is not among the samples.

Taking out the factor $\frac{P(G_M)}{p_{e'}} \frac{\partial p_{e'}}{\partial \theta_{e'}}$, we obtain
\begin{multline}
\frac{\partial z_i(v)}{\partial\theta_{e'}} \approx \frac{P(G_M)}{p_{e'}}\frac{\left( \binom{n}{2} + 1 \right)}{|\lbrace G \rbrace |} \frac{\partial p_{e'}}{\partial \theta_{e'}} \Bigg( \mathcal{K}(G_M,\mathcal{N}_{\mathcal{G}}^r(v)) + \\
\sum_{e \neq e'}\mathcal{K}(G_e,\mathcal{N}_{\mathcal{G}}^r(v))\frac{1-p_e}{p_e} - \mathcal{K}(G_{e'},\mathcal{N}_{\mathcal{G}}^r(v)) \Bigg)\,.
\end{multline}
Finally, we note that $\frac{\partial p_{e'}}{\partial \theta_{e'}} = \pm 1$, with the sign being positive when the edge is observed in the mode, negative when it is not.

Recall that, in our setup, each node of the graphs and masks has an associated distribution over node labels. During inference, we sample from these distributions to obtain the labeled graphs to pass to the kernel. During the learning phase, we compute the gradient of the output $z_i(v)$ \wrt the feature distribution of the mask by differentiating its expectation with a similar Monte Carlo approach. However, in this case, the expectation is taken not only over the structures of the graphs sampled from the mask, but also over the labels sampled from the corresponding distribution $\ell$ associated with each node in the mask
\red{\begin{equation}
z_i(v)=\mathbb{E}_{G\sim\mathcal{M}_i,\,l\in\ell^{\otimes n}}[\mathcal{K}(G^l,\mathcal{N}_{\mathcal{G}}^r(v))]
\end{equation}}

where $G^l$ is the graph with attached node labels $l$ and $\ell^{\otimes n}$ is the concatenated label distribution over all the nodes of the mask.

Assuming again that the distribution over the structure is concentrated around the mode $G_M$, we can then estimate the gradient \wrt the label distribution $\ell^{\otimes n}$
\begin{multline}
\nabla_{\ell^{\otimes n}} z_i(v) \approx \\
\frac{1}{|\{l\}|}\sum_{l \in \ell^{\otimes n}} (l-\bar{l})\bigg(\mathcal{K}(G_M^{l},\mathcal{N}_{\mathcal{G}}^r(v))-\mathcal{K}(G_M^{\bar{l}},\mathcal{N}_{\mathcal{G}}^r(v))\bigg)
\end{multline}
where $\bar{l}$ is the label set sampled in the forward pass, and $l$ and $\bar{l}$ taken as vectors are the concatenated one-hot encodings of the sampled labels.

Figure~\ref{fig:prob_evolution} shows the matrix of edge observation probabilities associated to a sample mask $\mathcal{M}_i$ for successive training epochs (top) and the corresponding mode graphs (bottom). Note that as the epoch number increases the probability distribution converges to the mode. This allows us to consider only the mode during inference, which results in a faster and deterministic prediction, \ie we do not need to estimate the expected value of the output of our model. Note that here, as well as in the experiments described in Section~\ref{sec:experiments}, we initialized the edge observation probabilities with random samples from a uniform distribution. Alternatively, one could initialize these probabilities to reflect the presence of frequent subgraphs mined from the training data or predefined patterns of interest, such as stars and cycles. We decide instead to limit the bias we introduce in the model, with an additional motivation being the fact that initializing the probabilities in this way appears to give the network sufficient flexibility to learn masks that resemble discriminative patterns found in the input graphs (see Section~\ref{sec:experiments} for further details).

\begin{figure}[t!]
\centering
\includegraphics[width=1\linewidth]{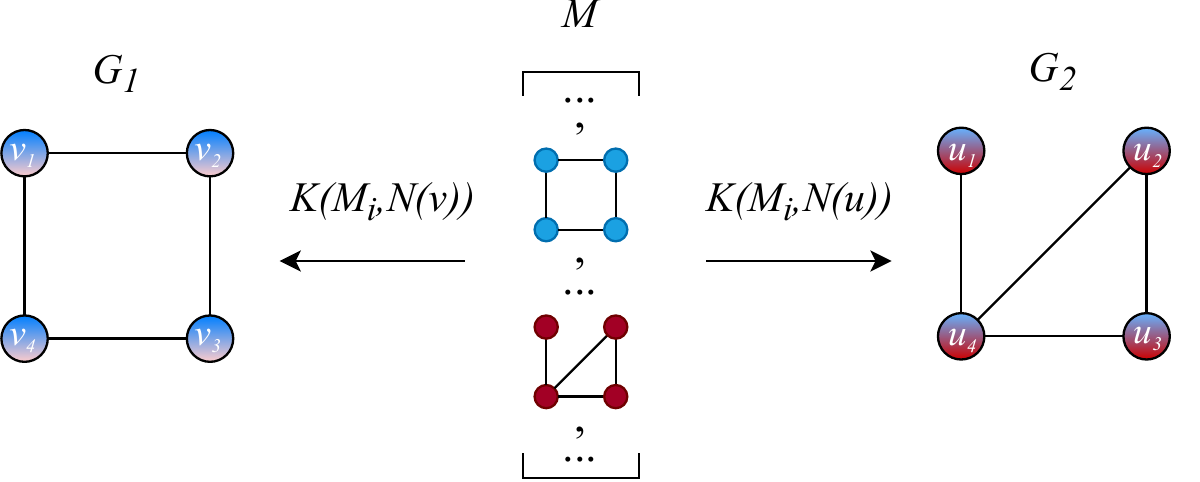}
\caption{Assuming the set of masks $M$ spans the space of graphs over $n$ nodes, if the radius of each $N_{\mathcal{G}_1}^r(v)$ is such that it spans the entire graph, the responses on the nodes of $\mathcal{G}_1$ will be identical and will peak for any mask fractionally equivalent to $\mathcal{G}_1$. Here, we use blue and red to highlight the masks with the highest response on $\mathcal{G}_1$ and $\mathcal{G}_2$, respectively.}
\label{fig:WL_success}
\end{figure}

\subsection{Expressive power}
In~\cite{morris2019weisfeiler}, Morris \etal show that the expressive power of standard GNNs (\ie GNNs that only consider the immediate neighbors of a node when updating the labels) is equivalent to that of the 1-WL test. In this subsection, we discuss our model's expressive power and argue that this is higher than the 1-WL test and thus standard message-passing GNNs. For this to be the case, our model should be both able to distinguish every pair of graphs that can be distinguished by the 1-WL test and it should be able to distinguish between pairs of graphs where the 1-WL test fails.
\\
\\
\noindent \textit{\textbf{Graphs distinguished by the 1-WL test.}}
Consider two unlabelled graphs $\mathcal{G}_1 = (V_1,E_1)$ and $\mathcal{G}_2 = (V_2,E_2)$ with the same number of nodes $n = |V_1| = |V_2|$, such that the 1-WL test can distinguish between them. Then we can also build a GKNN able to distinguish between them by adopting the WL kernel in the GKC layer, and the radius $r$ is such that the subgraphs $N_{\mathcal{G}_1}^r(v)$ span the entire graph $\mathcal{G}_1$ (similarly for $\mathcal{G}_2$). In fact, the set of labels computed against any one mask will be the same for every node in graph $\mathcal{G}_1$ ($\mathcal{G}_2$) since all the subgraphs are identically equal to the graph itself. Assuming that we have a set of structural masks $\lbrace \mathcal{M}_1, \cdots, \mathcal{M}_m\rbrace$ that span the space of graphs over $n$ nodes, then, under the hypothesis that the 1-WL test can distinguish between $\mathcal{G}_1$ and $\mathcal{G}_2$, the set of labels computed by the GKC layer (\ie the responses of the masks) while identical on all the nodes of each graph, will differ between the two graphs (see Figure~\ref{fig:WL_success}). In particular, the response will peak for any mask fractionally equivalent to the graph itself~\cite{ramana1994fractional}. Therefore, under any reasonable pooling strategy, the GKNN will be able to distinguish between the two graphs.

We can easily see that the same holds for an arbitrary subgraph radius $r$ under the assumption of a sufficient number of layers. For simplicity, we will assume $r=1$, but, as shown in the previous argument, increasing $r$ increases the descriptive power of the GKC layer. Note that the full GKC Layer can be seen as 1) a label propagation step followed by 2) a hashing step. The label propagation is given by the computation of the WL kernel against a structural mask, due to the 1-WL iterations on the subgraphs.

\begin{figure}[t!]
\centering
\includegraphics[width=1\linewidth]{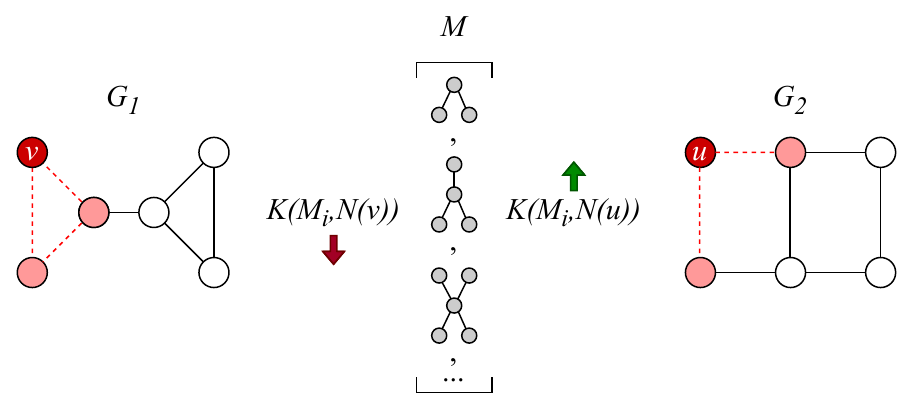}
\caption{$\mathcal{G}_1$ and $\mathcal{G}_2$ cannot be distinguished by the 1-WL test, however the GKNN correctly identifies them as non-isomorphic graphs. If $M$ is the set of all structural masks representing stars up to $n-1$ nodes, no star mask will achieve maximum similarity on the radius 1 subgraphs centered on the nodes of $\mathcal{G}_1$, while the same maximal response will be achieved when considering the radius 1 subgraphs centered on the nodes of $\mathcal{G}_2$.}
\label{fig:WL_failure}
\end{figure}

The hashing, on the other hand, comes from the implicit dot product against the structural masks. This is, in essence, a projection onto the space spanned by them and forces a hashing where the adopted masks form a representation space for the hashing itself. Note that, under the assumption of a sufficiently large (possibly complete) set of structural masks $\lbrace \mathcal{M}_1, \cdots, \mathcal{M}_m\rbrace$ spanning the space of graphs over $n$ nodes, the dot product induced hashing will not introduce conflicts between nodes with different labels. In other words, distinct labels under the 1-WL iterations will eventually be mapped to distinct labels at some GKC layer. Note, however, that under these assumptions the expressive power is greater or equal than that of 1-WL and labels that are equivalent under 1-WL iteration can indeed be mapped to distinct labels by some GKC layer. In fact, the propagation for 1-WL uses only the star of radius 1 around a particular node, while our scheme uses the full 1-hop subgraph, which includes connections between neighbors. Therefore, given a sufficient number of GKC layers and a reasonable pooling strategy, the GKNN will be able to distinguish between the two input graphs.
\\
\\
\noindent \textit{\textbf{Graphs where the 1-WL test fails.}}
Not only the GKNN can distinguish between pairs of graphs that are correctly distinguished by the 1-WL test, but it also succeeds where the 1-WL test fails. To see that this is the case, consider the standard example of two graphs with 6 nodes, where $\mathcal{G}_1$ is the dumbbell graph obtained by connecting two 3-nodes cycles with one edge, while $\mathcal{G}_2$ is a $2 \times 3$ lattice graph (see Figure~\ref{fig:WL_failure}). Both graphs have the same degree distribution, with 4 nodes of degree 2 and 2 nodes of degree 3. As explained in~\cite{sato2020survey,nikolentzos2020k}, despite being non-isomorphic, the two graphs are considered identical by the 1-WL test (in other words, the WL relabeling procedure converges to the same set of labels for the two graphs). To see why this happens, consider that through the propagation phases, each node is only made aware of what degree other nodes in the graph have, and not how many such nodes can be reached. However, we can easily see that this is not an issue for the GKNN.

\begin{table*}[t]\centering
\caption{Best models average training time per epoch (over folds) and inference time for the datasets considered in the present work.
}\label{tab:speed}
\begin{tabular}{l|cccc|cccc}\toprule
&\multicolumn{4}{c|}{Per epoch training time (s)} &\multicolumn{4}{c}{Per sample inference time (ms)} \\\cmidrule{2-9}
& Ours (WL) & GIN & DiffPool & ECC & Ours (WL) & GIN & DiffPool & ECC \\\midrule
PROTEINS &111.92 ± 59.85 &1.55 ± 0.14 &5.45 ± 0.20 &2.68 ± 0.21 &14.73 ± 5.48 &4.09 ± 0.94 &13.71 ± 0.16 &9.02 ± 1.29 \\
MUTAG &10.01 ± 1.92 &0.21 ± 0.04 &0.70 ± 0.03 &0.32 ± 0.05 &16.66 ± 1.12 &3.29 ± 1.18 &9.25 ± 0.18 &7.28 ± 1.03 \\
PTC &9.30 ± 3.67 &0.40 ± 0.07 &1.49 ± 0.10 &0.59 ± 0.05 &11.76 ± 4.68 &3.24 ± 1.22 &9.38 ± 0.19 &8.18 ± 0.67 \\
NCI1 &83.45 ± 1.80 &5.86 ± 0.72 &17.90 ± 0.52 &9.07 ± 0.44 &8.02 ± 0.36 &4.43 ± 0.69 &9.35 ± 0.16 &8.91 ± 1.71 \\
IMDB &36.80 ± 3.59 &1.44 ± 0.22 &4.25 ± 0.15 &1.88 ± 0.12 &8.90 ± 0.16 &4.23 ± 0.72 &9.25 ± 0.16 &7.64 ± 1.15 \\
ZINC &225.16 &6.21 &21.33 &12.45 &2.16 &4.33 &5.92 &4.80 \\
\bottomrule
\end{tabular}
\end{table*}

Assume we have a GKC layer with radius $r=1$ that adopts a WL kernel and a set of structural masks $M = \lbrace \mathcal{M}_1, \cdots, \mathcal{M}_m\rbrace$ representing all stars up to $n-1$ nodes, where $n$ denotes the number of nodes of the input graph. The presence of a triangle in the graph induces the existence of at least three subgraphs of radius 1 with connections between neighbors, \ie not stars. On these subgraphs, no star masks will achieve maximum similarity under the WL kernel. Having a pooling process that normalizes the masks responses \wrt size and maximizes over the masks on the same node, implies that each node will have the same maximal response if it does not contain triangles and a lower response otherwise. A min-pooling process over the node features will then be able to discriminate between graphs with and without triangles. Now consider the graphs $G_1$ and $G_2$ introduced above where the 1-WL test failed. As shown in Figure~\ref{fig:WL_failure} the subgraphs built around the nodes of $\mathcal{G}_1$ and $\mathcal{G}_2$ will be structurally different, \ie triangles in $\mathcal{G}_1$ and 3 nodes path graphs in $\mathcal{G}_2$. As a consequence, the feature vectors $x(v)$ over the nodes of the two graphs will be different, allowing the GKNN to distinguish between them.
\\
\\
\noindent \textit{\textbf{Empirical evaluation.}}
To further strengthen the argument that our network is able to distinguish graphs where the 1-WL test fails, we empirically evaluate the ability of a GKNN to distinguish between the two graphs of Figure~\ref{fig:WL_failure}. Specifically, we train a GKNN using the WL kernel with a single mask with a maximum number of nodes set to either 3 or 4. In both cases, when we inspect the learned mask we find that its mode corresponds to the 3-nodes cycle (triangle). Indeed, the presence of triangles is what distinguishes $G_2$ from $G_1$ and allows the GKNN to discriminate between the two graphs successfully.

\subsection{Computational complexity and runtime analysis}
The computational complexity of our model is $O(NK(s))$, where $N$ is the number of nodes of nodes of the input graph and $K(s)$ is the complexity of the kernel computation, which in turn depends on the size $s$ of the subgraphs. Note that $s \leq N$, where equality is attained for every subgraph in the case of complete graphs. More precisely, if we limit ourselves to the case of graphs with bounded degree, we have $s=O(\min{(N,d_{\max}^r)})$ where $d_{\max}$ denotes the maximum node degree.

In terms of runtime, for the datasets considered in this paper, this translates to the results shown in Table~\ref{tab:speed}, where we report the average epoch training time and the average inference time per dataset obtained during the grid search (for the best model, for each dataset and fold). Both training and testing of each model were performed using an Intel(R) Xeon(R) Silver 4114 CPU @ 2.20GHz processor and an NVIDIA Tesla V100. Note that in order to exploit the computational power of GPUs better we implemented a GPU version of the WL kernel. \red{Although the resulting inference runtime is higher than three widely used baselines, we compared our model to (ECC~\cite{simonovsky2017dynamic}, DiffPool~\cite{ying2018hierarchical}, and GIN~\cite{xu2018powerful}), it is of a similar order of magnitude.} When it comes to training, on the other hand, the computational bottleneck of our network is the computation of the numerical gradient, which in turn explains the subpar runtime.

\red{\subsection{Relation to existing works}
The closest works to ours we are aware of are~\cite{nikolentzos2020random,feng2022kergnns}. Nikolentzos \etal propose a GNN where the first layer consists of a series of hidden graphs that are compared against the input graph using a random walk kernel~\cite{nikolentzos2020random}. However, due to their optimization strategy, their model only works with a single differentiable kernel, whereas our model allows us to tap into the expressive power of any type of kernel. Moreover, in the architecture of~\cite{nikolentzos2020random}, the learned structural masks are assumed to be complete weighted graphs, whereas we allow for graphs with arbitrary structure. Finally, note that~\cite{feng2022kergnns} works very much in the same way of~\cite{nikolentzos2020random}, as it relies on the underlying graph kernel to be differentiable and, more specifically, it makes use of the random walk graph kernel. Another similarity with~\cite{nikolentzos2020random} is that the model proposed in~\cite{feng2022kergnns} is also not purely structural, as opposed to ours. Instead, it learns matrices of weights which, in order to be interpreted as learned structural patterns, need to be first thresholded using a ReLU function that prunes edges associated with smaller weights.}
\section{Experimental evaluation}\label{sec:experiments}
In this section we analyze the model hyper-parameters, evaluate the proposed architecture on graph classification and regression tasks, and compare it with widely used baselines and other GNN models. A Python implementation of our model is available at \url{https://github.com/gdl-unive/Graph-Kernel-Convolution}.

\subsection{Datasets}\label{sec:datasetdetails}
In our experiments we make use of 6 graph classification and 1 graph regression real-world datasets.

MUTAG~\cite{debnath1991structure} is a dataset consisting of 188 mutagenic aromatic and heteroaromatic nitro compounds (with 7 discrete labels) assayed for mutagenicity in Salmonella typhimurium. The goal is predicting whether each compound possesses mutagenicity effect on the Gram-negative bacterium Salmonella typhimurium. 
 
NCI1 (the anti-cancer activity prediction dataset) is a balanced subset of datasets of chemical compounds screened for activity against non-small cell lung cancer and ovarian cancer cell lines~\cite{wale2008comparison}. It has 37 discrete labels.
 
The PROTEINS~\cite{borgwardt2005protein,dobson2003distinguishing} dataset consists of proteins represented as graphs.
Nodes represent the amino acids and an edge occurs between two vertices if they are neighbors in the amino-acid sequence or 3D space. It has 3 discrete labels, representing helix, sheet, or turn. The aim is to distinguish proteins into enzymes and non-enzymes. 
 
The PTC (Predictive Toxicology Challenge) dataset~\cite{helma2001predictive,kriege2012subgraph} is a collection of chemical compounds reporting the carcinogenicity for  Male Rats (MR), Female Rats (FR), Male Mice (MM), and Female Mice (FM). We selected graphs of Male Rats (MR) for evaluation, consisting of 344 graphs very small and sparse with 19 discrete labels. 

ENZYMES consists of protein tertiary structures obtained from~\cite{borgwardt2005protein}, in particular 600 enzymes from the BRENDA enzyme database~\cite{schomburg2004brenda}. The task for this dataset is assigning each enzyme to one of the 6 EC top-level classes. For our experiments, we considered only 3 discrete labels.

\begin{figure}[t!]
\centering
\begin{overpic} [trim=-0.8cm -1.1cm 0cm -0.5cm, clip, width=0.34\linewidth]{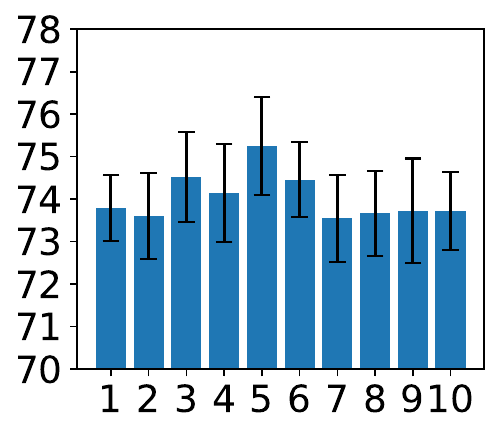}
\put(0,28){\rotatebox[origin=lB]{90}{\footnotesize \textit{Accuracy (\%)}}}
\put(57,92.5){\makebox(0,0){\footnotesize \textit{\# Nodes}}}
 \end{overpic}
\begin{overpic} [trim=0cm -1.1cm 0cm -0.5cm, clip, width=0.31\linewidth]{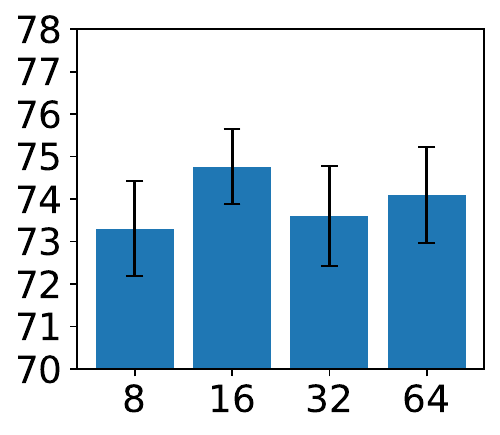}
\put(55,96){\makebox(0,0){\footnotesize \textit{\#Structural masks}}}
 \end{overpic}
\begin{overpic} [trim=0cm -0.3cm -0cm -0.5cm, clip, width=0.31\linewidth]{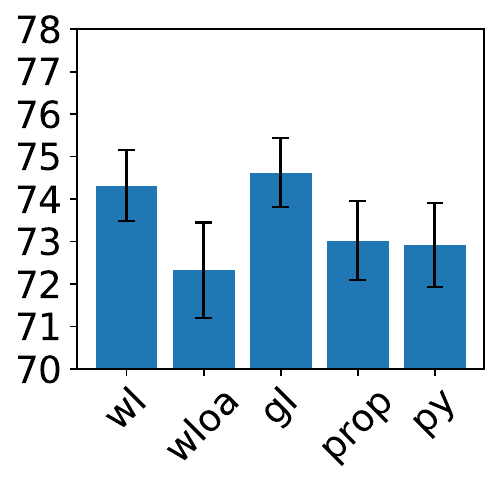}
\put(57,94){\makebox(0,0){\footnotesize \textit{Graph kernel}}}
\end{overpic}
\begin{overpic} [trim=-0.8cm -0.9cm -0cm -0.5cm, clip, width=0.34\linewidth]{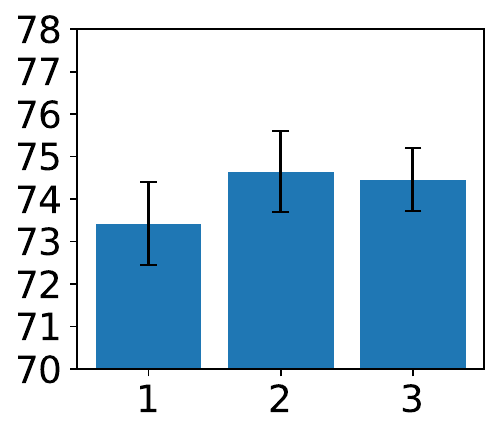}
\put(0,28){\rotatebox[origin=lB]{90}{\footnotesize \textit{Accuracy (\%)}}}
\put(60,90.5){\makebox(0,0){\footnotesize \textit{Radius $r$}}}
\end{overpic}
\begin{overpic} [trim=-0.1cm -1.0cm -0.0cm -0.5cm, clip, width=0.31\linewidth]{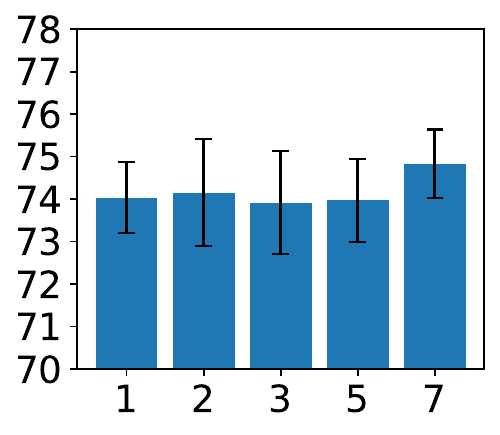}
\put(53,94.5){\makebox(0,0){\footnotesize \textit{\# Layers}}}
\end{overpic}
\begin{overpic} [trim=-0.1cm 0.2cm -0.1cm -0.5cm, clip, width=0.31\linewidth]{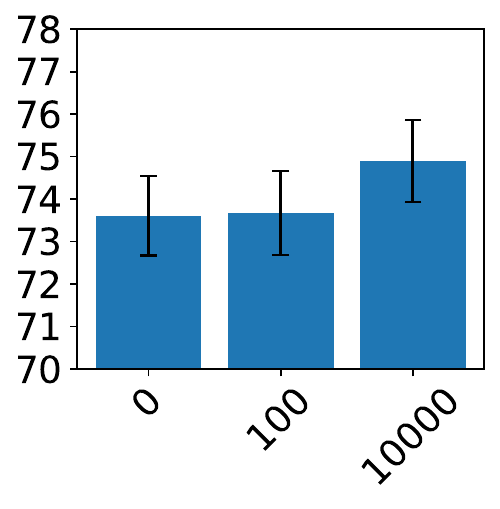}
\put(55,94.5){\makebox(0,0){\footnotesize \textit{JSD weight}}}
\end{overpic}
\vskip -2mm
\caption{Ablation study: bar plots of classification accuracy with standard error. Left to right, top to bottom: number of nodes (\ie structural mask size), number of structural masks, kernel functions (WL, WL with Optimal Assignment, Graphlet, Propagation, and Pyramid match kernel), subgraph radius, number of GKC layers, and weight of the JSD loss.}
\label{fig:ablation}
\end{figure}

The IMDB-BINARY/MULTI datasets~\cite{yanardag2015deep} collect graphs about movie collaboration. Each graph consists of nodes representing actors/actresses who played roles in movies in IMDB, and an edge between two vertices takes place if they appear in the same movie. Graphs are derived from the Action and Romance genres. The task is to identify which genre an ego-network graph belongs to. Note that while the bio/chemo-informatics graphs described above have categorical features, the nodes have no features in the IMDB-BINARY/MULTI datasets.

For the graph regression task, we employ a common benchmark molecular dataset, ZINC~\cite{irwin2012zinc}, a drug-constrained solubility prediction dataset. In particular, we use a subset (12K) of the ZINC molecular graphs (250K), as in~\cite{dwivedi2020benchmarking}. Here, the goal is regressing the constrained solubility, which is the desired molecular property. The node features in each molecular graph refer to the types of heavy atoms. Even though the dataset takes into consideration the types of bonds between nodes encoded as edge features, we do not use them in our experiments.

In order to explore the ability of our model to provide an insight on the structural patterns being learned, we also built 5 further synthetic datasets based on 5 graph motifs. For more information on these datasets see Subsection~\ref{sec:interpretability}.

Finally, note that in this work we restrict our attention to datasets composed of undirected graphs with categorical node attributes. However, in principle, the proposed GKNN can also operate with graph kernels that allow for directed, edge-attributed graphs. Continuous features can also be dealt with as long as the kernel is differentiable \wrt them.

\begin{figure}[t!]
\centering
\begin{overpic} [trim=-0.6cm 0cm 0cm 0cm, clip, width=0.9\linewidth]{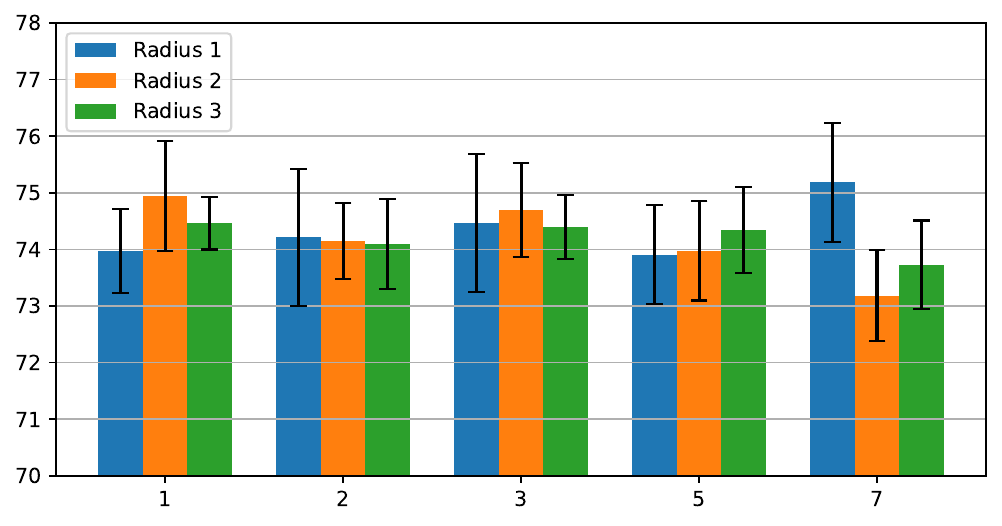}
\put(0,16){\rotatebox[origin=lB]{90}{\footnotesize \textit{Accuracy (\%)}}}
\put(52,0){\makebox(0,0){\footnotesize \textit{\# Layers}}}
 \end{overpic}
\caption{Ablation study: bar plots of classification accuracy with standard error. Each group of bars shows the accuracy of models with a given depth, while the different colors within each group correspond to different subgraph radii.}
\label{fig:ablation_lvsh}
\end{figure}

\begin{figure}[t!]
\centering
\begin{overpic} [trim=-7mm -4mm 2mm 0, clip, width=0.495\linewidth]{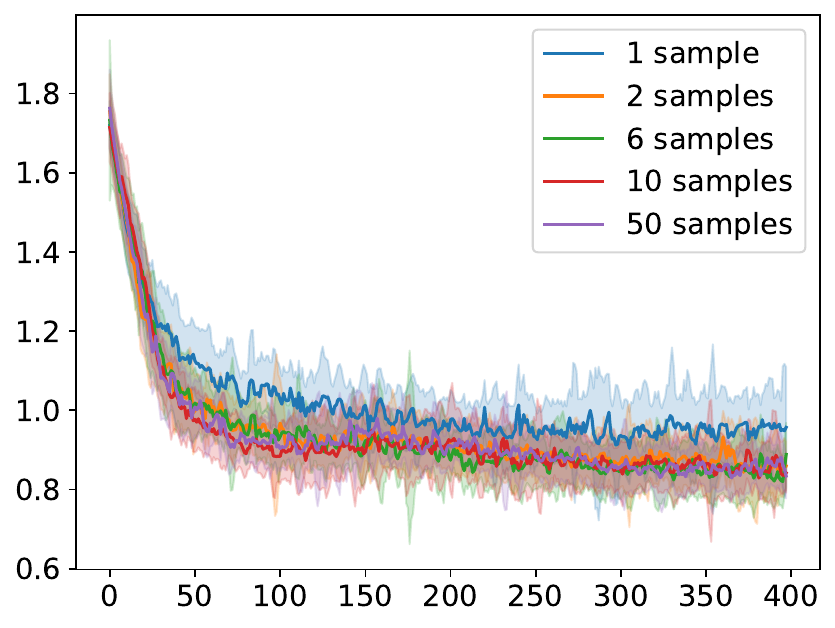}
\put(60,0){\makebox[0pt]{\scriptsize Epochs}}
\put(0,18){\rotatebox{90}{\scriptsize Cross-entropy loss}}
\end{overpic}
\begin{overpic} [trim=-3mm -4mm 2mm 0cm, clip, width=0.491\linewidth]{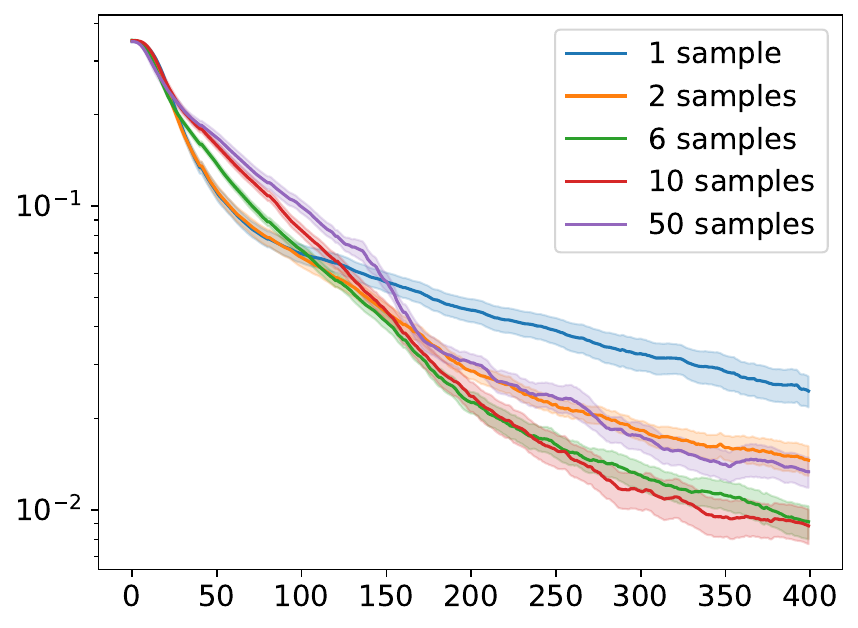}
\put(60,0){\makebox[0pt]{\scriptsize Epochs}}
\put(0,30){\rotatebox{90}{\scriptsize Masks entropy}}
\end{overpic}
 \caption{Cross-entropy loss (left) and masks entropy (right) for increasing number of samples during the important sampling optimization. Each model has been trained with 8 masks for 400 epochs on the first fold of MUTAG. We report curves with standard error computed on 20 runs for each number of samples.}
\label{fig:num_samples}
\end{figure}

\subsection{Ablation study}\label{sec:ablation}
We start by running an ablation study on the PROTEINS dataset, where we investigate how the various components and hyper-parameters of our architecture affect the model performance. To this end, we perform a grid search over all the studied hyperparameter values where we compute the accuracy for a specific hyperparameter and value pair by averaging the results over the 3 best performing models for each fold. Specifically, we study the influence of the number of nodes (\ie the maximum size of the structural masks), number of structural masks, kernel function, subgraph radius, number of GKC layers, and weight of $loss_{JSD}$.

\begin{figure}[t!]
\centering
\begin{overpic} [trim=-2.2cm 0cm 0cm -2cm, clip, width=1\linewidth]{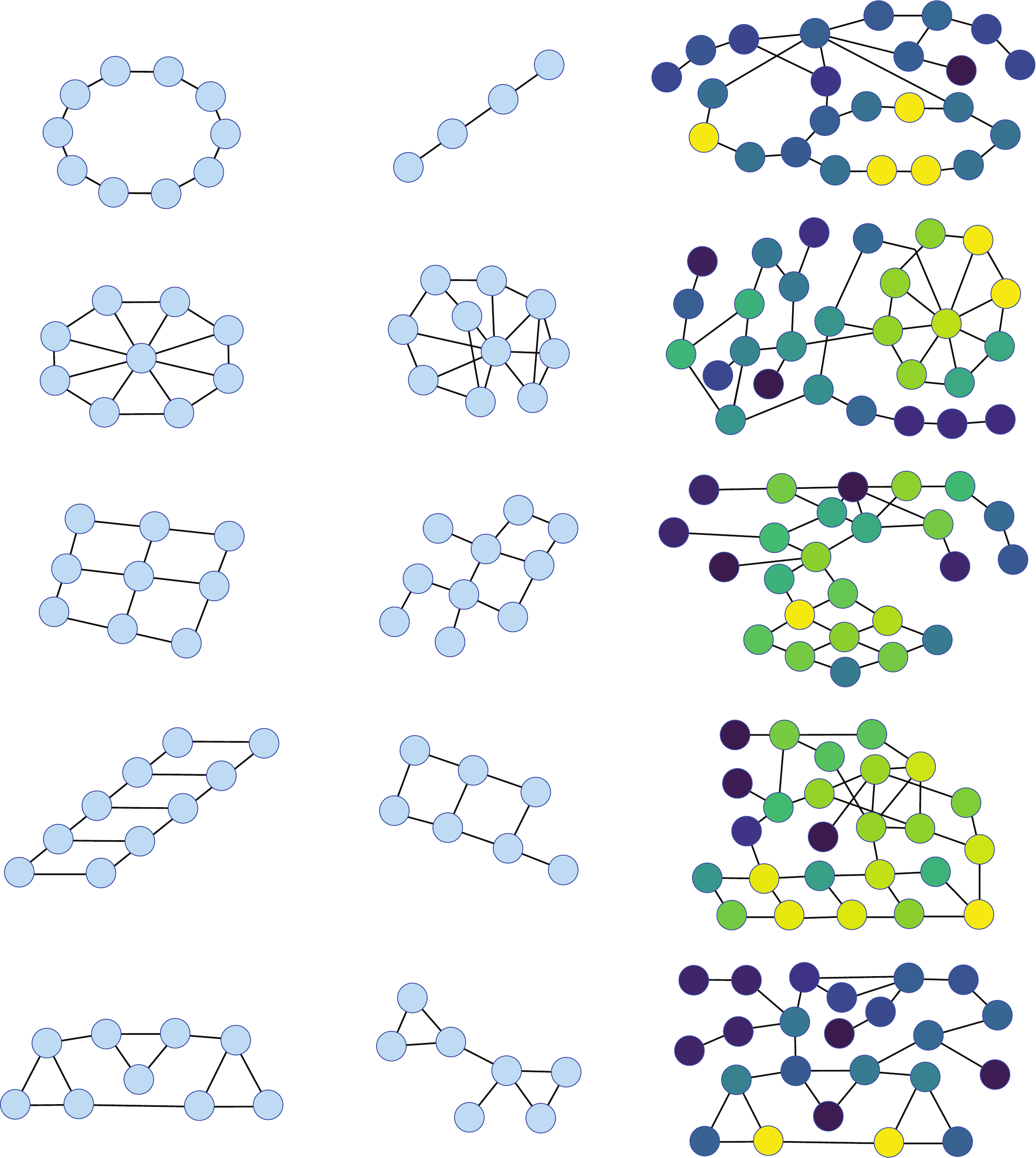}
\put(0,83){\rotatebox{90}{\textit{Ring}}}
\put(0,62){\rotatebox{90}{\textit{Wheel}}}
\put(0,44){\rotatebox{90}{\textit{Grid}}}
\put(0,23){\rotatebox{90}{\textit{Ladder}}}
\put(0,2){\rotatebox{90}{\textit{Cliques}}}

\put(15,98){\makebox(0,0){\textit{Motif}}}
\put(42,98){\makebox(0,0){\textit{Structural Mask}}}
\put(74,98){\makebox(0,0){\textit{Node response}}}
\end{overpic}

 \vskip 0mm
 \caption{Interpretability analysis results. Each row refers to a graph motif. Left: original motif. Mid: structural mask with the strongest response. Right: a sample graph including the motif. Colors indicate the node response to the filter, with lighter colors (yellow) indicating a high response and darker colors (blue) indicating a low response.}
\label{fig:structures}
\end{figure}

We report the average accuracy results as bar plots with standard error in Figure~\ref{fig:ablation} (from left to right: number of nodes, number of structural masks,  kernel function, subgraph radius, number of GKC layers, and weight of the JSD loss). Both the number of structural masks and the maximum number of their nodes play an important role in the classification accuracy and thus need to be carefully chosen. Note that we only fix the maximum number of nodes produced by a given mask, meaning that, in principle, at convergence, the model may have learned masks corresponding to graphs with fewer nodes. Also, while nothing in our framework prevents us from fixing a different maximum for each mask, for simplicity and to reduce the number of free parameters, we set the maximum to be the same for all masks.

As expected, the choice of the kernel function is also a crucial factor, highlighting the advantage of allowing plug-in any non-differentiable graph kernel function. Indeed, one can imagine that the choice of the kernel affects the type of structural patterns our network can capture and, therefore, impacts its overall performance. For instance, if we were to choose a kernel that does not distinguish between the presence and absence of a given structural pattern, and if this pattern was important to discriminate between two classes of graphs, we can foresee the performance of our network to be negatively impacted.

As for the subgraph radius, the number of neighbors clearly influences the model performance, and there seems to be an optimal subgraph size that is able to catch the structural characteristics of the input graphs. The results also confirm the importance of the JSD loss as the performance of the GKNN increases with its use.

\begin{figure}[t!]
    \centering
    \begin{overpic} [ trim=0cm -0cm 0.0cm -1.5cm, clip, width=1\linewidth]{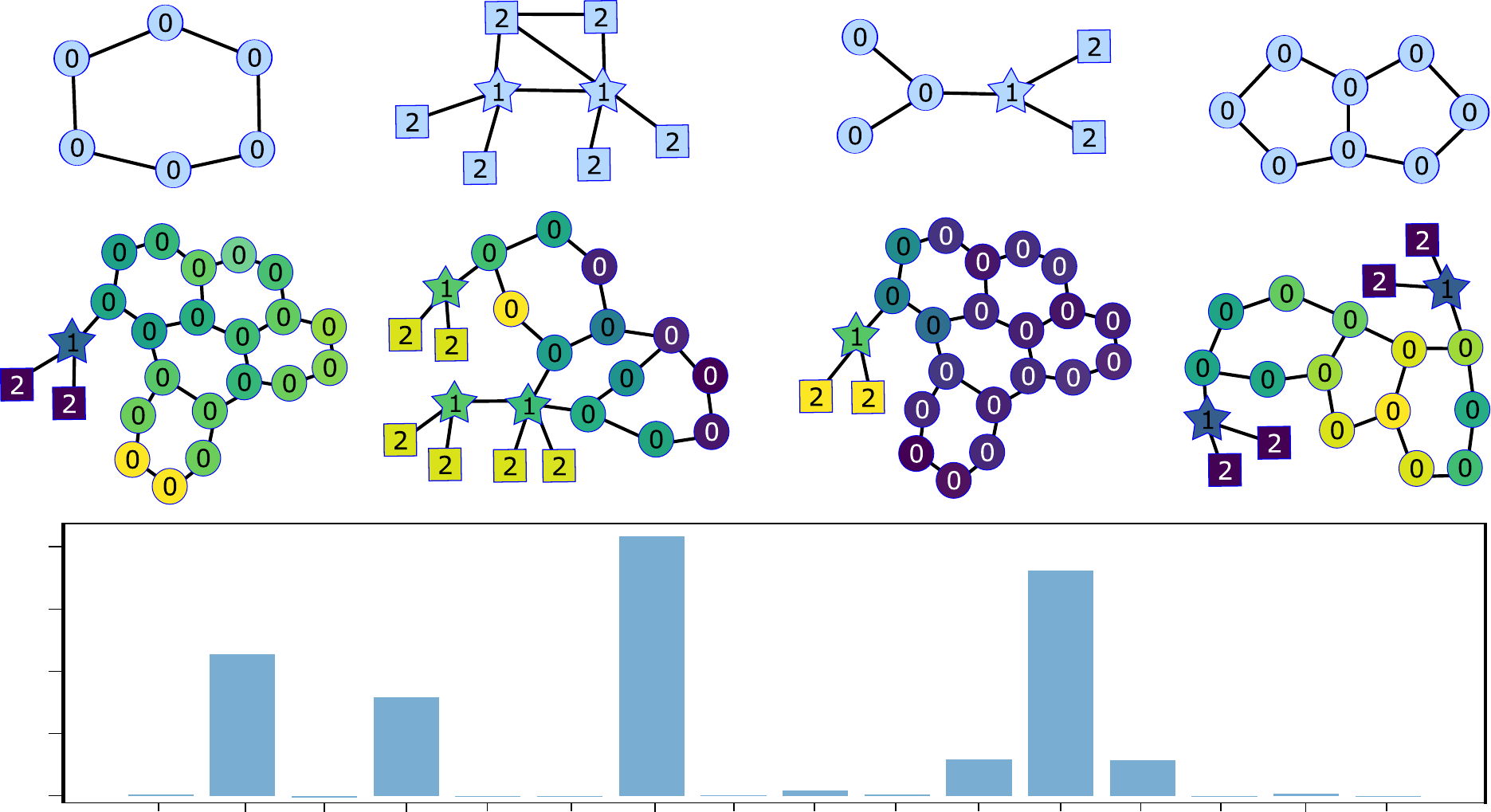}
    \put(0,1){\scriptsize\rotatebox{90}{Loss increase}}
    \put(9,-3){\tiny $\gM_1$}
    \put(90,-3){\tiny $\gM_{16}$}
    \put(45,-2.8){\scriptsize $\cdot$}
    \put(55,-2.8){\scriptsize $\cdot$}
    \put(65,-2.8){\scriptsize $\cdot$}
    \put(8,56){\footnotesize $\gM_2$}
    \put(34,56){\footnotesize $\gM_4$}
    \put(63,56){\footnotesize $\gM_7$}
    \put(88,56){\footnotesize $\gM_{12}$}
     \end{overpic}
     \vskip 1mm
    \caption{Top 4 significant structural masks (top) learned by our model on the MUTAG dataset and their response over input graph nodes (middle). Shape and number both represent the node label. The bottom bar plot shows the impact of each structural mask on the classification loss.}
    \label{fig:MUTAG_masks}
\end{figure}

We also study the relation between the subgraphs radius and the number of layers of the model. The results of Figure~\ref{fig:ablation_lvsh} show that the optimal number of layers is connected to the subgraphs radius. The models with radius 1 need more layers to perform better, while wider subgraphs tend to decrease the performance with deeper networks. Indeed, when using smaller subgraphs, more convolution layers are needed to collect information from a larger neighborhood.

Finally, Figure~\ref{fig:num_samples} shows the influence of the number of samples drawn during the importance sampling (see Section~\ref{sec:importance}) in terms of model performance (left) and entropy of the masks (right). The results suggest that our model is robust \wrt the choice of the sample size. Indeed, when the sample size is at least 2, its impact on the model performance appears to be negligible. At the same time, Figure~\ref{fig:num_samples} (right) shows that all masks converge to a low value of the entropy, regardless of the sample size. Overall, these results suggest that we can minimize the runtime by drawing a small number of samples without compromising the model performance accuracy. Accordingly, in our experiments, we fixed the sample size to 3.

\subsection{Interpretability analysis}
\label{sec:interpretability}
One of the factors that fostered interpretability in classical CNNs is the possibility to visualize and analyze the learned filters, capturing the fundamental structures characterizing the input images~\cite{nikolentzos2020random}. This feature is missing in the classical formulation of graph convolution relying on the message-passing paradigm. On the other hand, our method learns actual graph masks, potentially increasing the interpretability of the model by allowing us to probe into the explanatory factors of variation behind the input data through the learned structural masks.

To show the potential of our model to capture fundamental structures in the input graphs, we devised a simple but effective synthetic experiment. First, we train our model on a binary classification task, \ie predicting if a certain graph motif is present or not in the given graph. Then, we qualitatively assess whether the learned structural masks have captured the graph motif. Following the setup of~\cite{nikolentzos2020random}, we created 5 different datasets, one for each of the graph motifs depicted in Figure~\ref{fig:structures} (leftmost column). Each dataset is composed of 2,000 graphs, equally divided into positive and negative samples, with a varying number of nodes between 30 and 50. Each sample pair is generated starting from the same Erd{\H o}s–Rényi graph, with an edge observation probability of 0.1. Next, two new graphs are built for each Erd{\H o}s–Rényi graph. The first one, labeled as 1, is created by connecting each node of the Erd{\H o}s–Rényi to the nodes of the structural motif with probability 0.02. The second one, labeled as 0, is obtained in a similar manner; however, instead of a motif, we insert a random graph with the same number of nodes and edges of the motif structure. With these datasets to hand, the model is then trained end-to-end to predict the graph labels on a 90/10$\%$ train/test split. Note that we trained the network with the same hyper-parameters as in the ablation baseline in section~\ref{sec:ablation}, with the exception of the number of structural masks, which is set to 8.

The results show that the learned structural masks (\ie the mode of the learned edge and label probabilities) have very similar structures to those of the corresponding motifs, with only a few missing or misplaced edges (see for instance the columns corresponding to the ring and the wheel). Moreover, the distribution of the responses in the original graphs clearly highlights the motifs position. Overall, the results demonstrate that our model is able to extract the salient structural patterns, which in turn allows us to understand what features were detected for a given input graph. We would like to stress that even though this aspect is not the final goal of this work, the interpretability of our model definitely helps to foster trust in our approach.

To further investigate the ability of our model to capture salient structural features, we show in Figure~\ref{fig:MUTAG_masks} some of the structural masks learned on the MUTAG dataset. In particular, we trained a model with the hyper-parameters of the baseline architecture used in the ablation study of Section~\ref{sec:ablation}.
To select the most significant filters, we manually set to zero the response of the $i$th filter $\gM_i$ and evaluate again the classification loss. In the bar plot (Figure~\ref{fig:MUTAG_masks}, bottom), we report the loss increase after zeroing each filter response. Below each mask, we also show the input graph of the training set that gave the highest kernel response among all nodes, together with the per-node response as node colors (high response in yellow, low response in blue). In summary, our analysis suggests that the learned masks can, to some extent, be interpreted as discriminative structural patterns that allow our network to distinguish between different classes of graphs.

\begin{figure}[t!]
\centering
\begin{overpic} [ trim=-1.75cm 0cm -5mm -1.5cm, clip, width=1\linewidth]{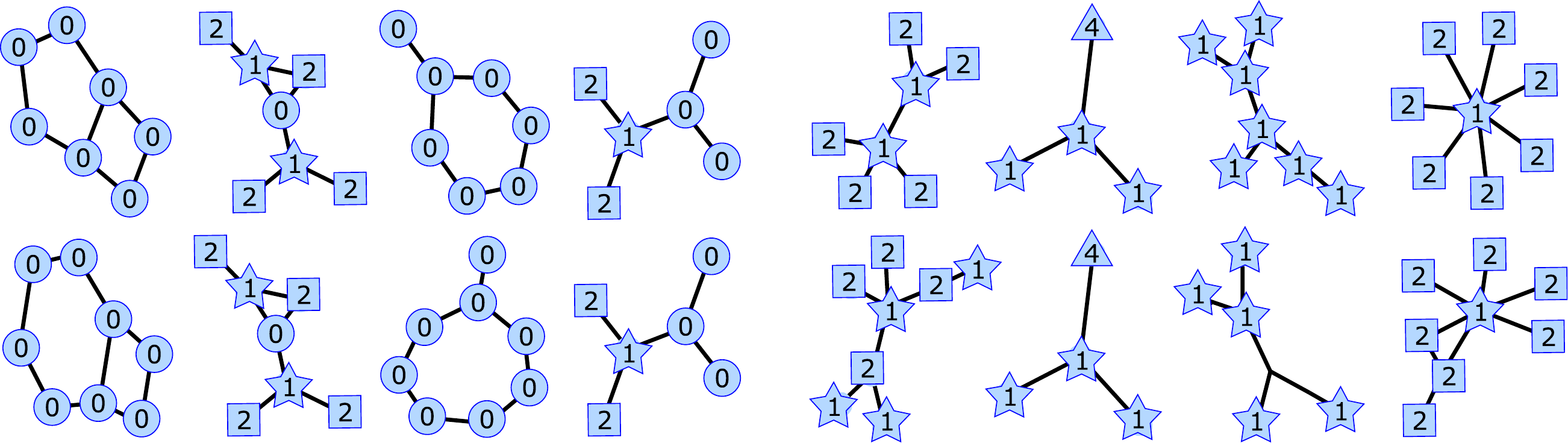}
\put(0,1){\scriptsize \rotatebox{90}{Model 2}}
\put(0,15){\scriptsize \rotatebox{90}{Model 1}}
\put(24,28){\scriptsize MUTAG}
\put(75,28){\scriptsize PTC}

\put(0.5,13.2){\line(1,0){100}}
\put(2.5,0){\line(0,1){27}}
\put(51,0){\line(0,1){27}}
\put(100,0){\line(0,1){27}}

 \end{overpic}
 \vskip -1mm
\caption{Top 4 significant structural masks learned by our model on the MUTAG (left) and PTC (right) datasets starting from two different random initializations of the model weights (rows). Shape and number both represent the node label.}
\label{fig:conv_masks}
\end{figure}

\begin{figure}[t!]
\centering
\begin{overpic} [ trim=-0cm -0cm -0cm -0cm, clip,  width=1\linewidth]{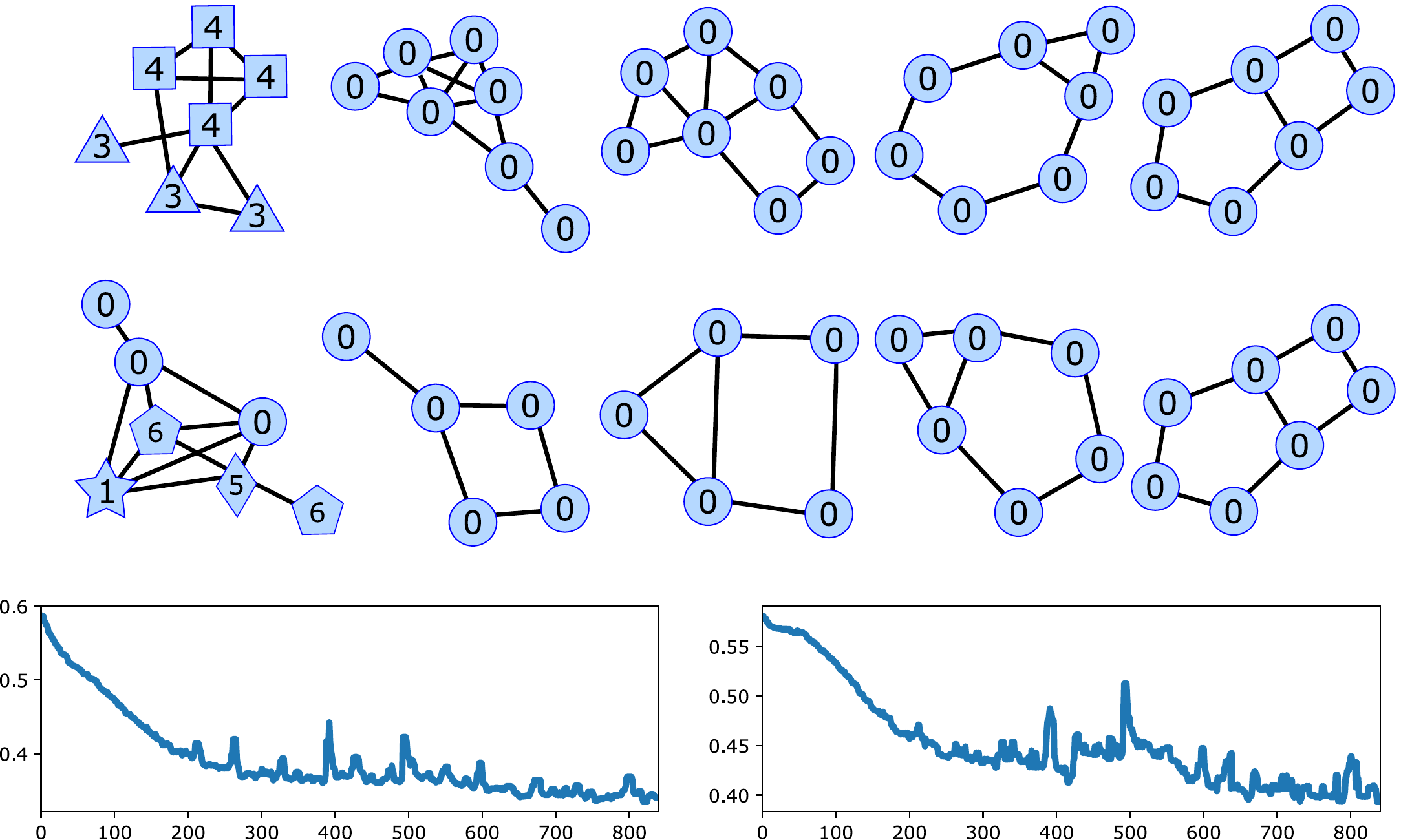}
\put(0,26){\scriptsize \rotatebox{90}{Model 2}}
\put(0,46){\scriptsize \rotatebox{90}{Model 1}}

\put(8,19){\tiny Initialization}
\put(30,19){\tiny Epoch 5}
\put(48,19){\tiny Epoch 110}
\put(66.5,19){\tiny Epoch 150}
\put(88,19){\tiny Final}

\put(8,40.7){\tiny Initialization}
\put(30,40.7){\tiny Epoch 5}
\put(48,40.7){\tiny Epoch 30}
\put(66.5,40.7){\tiny Epoch 160}
\put(88,40.7){\tiny Final}

\put(86,0){\tiny epochs}
\put(26,14){\tiny Train cross-entropy}
\put(72,14){\tiny Validation cross-entropy}

\put(0,40){\line(1,0){100}}
\put(3,21){\line(0,1){38}}
\end{overpic}
\caption{Top: Evolution of the learned structural masks during the training epochs showing two different initializations (belonging to two different models) that led to the same structural mask on the MUTAG dataset. Bottom: Cross-entropy loss during training.}    \label{fig:evolution}
\end{figure}

\subsection{Convergence analysis}\label{sec:convergence}
To experimentally assess the convergence of our optimization strategy, we investigated the behavior of the structural masks both at convergence and during the optimization. In Figure~\ref{fig:conv_masks}, we show the most significant (see~\ref{sec:interpretability} for the definition) structural masks learned by two different training instances of the same architecture. Both models end up learning similar structural masks, indicating that our optimization is able to reach a stable local optimum. This is also supported by the interpretability analysis of Section~\ref{sec:interpretability}, showing that the learned masks resemble the local sub-structures of the input graphs.

In Figure~\ref{fig:evolution}, we also investigate the evolution of different initializations of the structural mask weights leading to the same final structure in two different models trained on MUTAG. Node labels have a faster convergence than connectivity, which is probably due to their bigger impact on the similarity score computed by graph kernels (WL in this specific case). The bottom of the figure also reports the decreasing curves of the cross-entropy losses during training on both train and validation sets.

\begin{table*}[t!]\centering
\caption{Classification results on chemo/bio-informatics and social datasets. Mean accuracy (acc) or Mean Absolute Error (mae) and standard error are reported. The best performance (per dataset) is highlighted in bold, and the second best is underlined.}\label{tab:comparisons}
{
\centering
\begin{tabular*}{\textwidth}{@{\extracolsep{\fill}} l | cccc||cc || c}
&MUTAG (acc) &PTC (acc) &NCI1 (acc) &PROTEINS (acc) &IMDB-B (acc)&IMDB-M (acc) & ZINC (mae) \\
\hline
Baseline &78.57 ± 4.00 &58.34 ± 2.02 & 68.50 ± 0.87 &73.05 ± 0.90 & 49.50 ± 0.79 & 32.00 ± 1.17 & 0.656  ± 0.009\\
WL &82.67 ± 2.22 &55.39 ± 1.27 &\textbf{79.32 ± 1.48} & 74.16 ± 0.38 & \textbf{71.80 ± 1.03} & 43.33 ±  0.56  & 1.244  ± 0.015\\
\hline
DiffPool &81.35 ± 1.86 &55.87 ± 2.73 &75.72 ± 0.79 &73.13 ± 1.49 & 67.80 ± 1.44 & 44.93 ± 1.02 & 0.487 ± 0.007\\
GIN &78.13 ± 2.88 &56.72 ± 2.66 & \underline{78.63 ± 0.82} &70.98 ± 1.61 & 71.10 ± 1.65 & 46.93 ± 1.06 & \underline{0.381 ± 0.007}\\
DGCNN & 85.06 ± 2.50 &53.50 ± 2.71 &76.56 ± 0.93 & 74.31 ± 1.03 & 53.00  ± 1.32 & 41.80 ± 1.39 & 0.598  ± 0.005\\
ECC & 79.68 ± 3.78 & 54.43 ± 2.74 &    73.48 ± 0.71    & 73.76 ± 1.60 &  68.30 ± 1.56 &    44.33 ± 1.59  & 0.726  ± 0.011 \\
GraphSAGE &77.57 ± 4.22 &\underline{59.87 ± 1.91 }&75.89 ± 0.96 &73.11 ± 1.27 & 68.80  ± 2.26 & 47.20 ± 1.18 &  0.462  ± 0.007\\
sGIN &84.09 ± 1.72 &56.37 ± 2.28 &77.54 ± 1.00 &73.59 ± 1.47 & \underline{71.30 ± 1.75} & \textbf{48.60 ± 1.48} &  -\\
\hline
KerGNN & 82.43 ± 2.73 & 53.15 ± 1.83 & 74.16 ± 3.36 & 72.96 ± 1.46 & 67.90 ± 1.33 & 46.87 ± 1.43 & 0.568  ± 0.007\\
RWGNN &82.51 ± 2.47 &55.47 ± 2.70 &72.94 ± 1.16 &73.95 ± 1.32&69.90 ± 1.32 & 46.20  ± 1.08 & 0.547  ± 0.004\\
\textbf{Ours (WL)} &\textbf{85.73 ± 2.70} & 59.29 ± 2.54 & 73.63 ± 1.32 &\underline{74.94 ± 1.10} &69.70 ± 2.20 & \underline{47.87 ± 1.78} & \textbf{0.364  ± 0.005}\\
\textbf{Ours (GL)} &\underline{85.24 ± 2.28} & \textbf{60.13 ± 1.94} & 71.51 ± 1.20 &\textbf{75.36 ± 1.12} & 69.90 ± 1.44 &   45.67 ± 1.22  & - \\
\hline
\end{tabular*}}
\end{table*}

\subsection{Graph classification and regression}
\label{sec:classification_results}
We compare our model against: \begin{itemize*}
\item 6 state-of-the-art GNNs: ECC~\cite{simonovsky2017dynamic}, DGCNN~\cite{zhang2018end}, DiffPool~\cite{ying2018hierarchical}, GIN~\cite{xu2018powerful}, (s)GIN~\cite{di2020mutual}, and GraphSAGE~\cite{hamilton2017inductive};

\item two distinct baselines, depending on the dataset type (see below): Molecular Fingerprint~\cite{ralaivola2005graph,luzhnica2019graph} and Deep Multisets~\cite{zaheer2017deep};  

\item  the WL kernel~\cite{shervashidze2011weisfeiler}; 

\item RWNN~\cite{nikolentzos2020random} and KerGNN~\cite{feng2022kergnns}, two GNN models employing differentiable graph kernels, the closest existing neural architectures to ours.
\end{itemize*}
We use a $C$-SVM~\cite{chang2011libsvm} classifier for the WL subtree kernel. For the baselines, we follow~\cite{errica_fair_2020} and use the Molecular Fingerprint technique~\cite{ralaivola2005graph,luzhnica2019graph} for chemical datasets. For the social dataset, we rely on the permutation invariant model of~\cite{zaheer2017deep}.

To ensure a fair comparison, we follow the same experimental protocol for each of these methods. We perform 10-fold cross validation where in each fold the training set is further subdivided in training and validation with a ratio of 9:1. The validation set is used for both early stopping and to select the best model within each fold. Importantly, folds and train/validation/test splits are consistent among all the methods. \red{Note that in our experimental setup, the best performing model on the validation set is directly evaluated on the test set without first re-training it on the union of the training and validation sets.}

We perform a grid search for all the methods to optimize the hyper-parameters. In particular, for the WL method, we optimize the value of $C$ and the number of WL iterations $h \in \{4, 5, 6, 7\}$. For the RWGNN, we investigate the hyper-parameter ranges used by the authors~\cite{nikolentzos2020random}, while for all the other GNNs \red{and the baselines}, we follow~\cite{errica_fair_2020} and we perform a full search over the hyper-parameters grid. For our model, we explore the following hyper-parameters: number of structural masks in $\{8, 16, 32\}$, maximum number of nodes of a structural mask in $\{6, 8, 10\}$, subgraph radius in $\{1, 2, 3\}$, and number of layers in $\{1, 2, 3\}$. We implemented two versions of our network, one that makes use of the WL kernel~\cite{shervashidze2011weisfeiler} and one that makes use of the Graphlet (GL) kernel~\cite{shervashidze2009efficient} (the top 2 performing kernels identified in the ablation study). For the GL kernel, we use graphlets of size 5 on all datasets except for IMDB-B and IMDB-M, where we limit the size to 3\footnote{This is due to the presence of high degree nodes in the IMDB datasets and the fact that the computational complexity of the GL kernel is $O(Nd_{\max}^{k-1})$, where $N$ is the number of nodes of the graph and $d_{\max}$ is the maximum degree.}.

In all our experiments, we train \red{our} model for $1000$ epochs, using the Adam optimizer with a learning rate of $0.001$ for MLP weights and $0.01$ for the edit operation probabilities and a batch size of 32. The MLP takes as input the sum pooling of the node features and is composed by two layers of output dimension $m$ and $c$ (\# classes) with a ReLU activation in-between. We use cross-entropy and mean absolute error (MAE) losses for the graph classification and regression tasks, respectively.

\textbf{\textit{Results.}} The accuracy for each method and dataset is reported in Table \ref{tab:comparisons}. The performance of our approach \red{outperforms the baselines~\cite{ralaivola2005graph,zaheer2017deep,luzhnica2019graph} on all the datasets, with the improvement being statistically significant on all datasets except PTC. Compared to other GNNs,} the performance of our model is on par with the best ones, as we can see for MUTAG, PROTEINS, and PTC. This holds as long as the number of node labels is small. Indeed, in the case of high-dimensional node labels (\eg NCI1), the classification accuracy tends to decrease as the optimization becomes harder. As for the social dataset, the performance of our approach is satisfactory, particularly considering the lack of discriminative sub-structures in social networks. Finally, our method achieves the best performance (\ie lowest MAE) on ZINC, with only GIN achieving a similar yet higher MAE. It should also be noted that our model always outperforms or performs comparably to both KerGNN and RWGNN, the only two other models that allow some degree of interpretability of the learned structures.
\section{Conclusion}\label{sec:conclusion}
We introduced a new convolution operation on graphs based on graph kernels. We proposed the graph kernel convolution layer and an architecture that makes use of this layer for graph classification purposes. The benefits of this architecture include the definition of a fully structural model that can exploit the vast collection of existing graph kernels, a superior expressive power when compared to standard GNNs, and the possibility to visualize the learned substructures in a way that is reminiscent of the convolutional filters of standard CNNs. Future work will attempt to address the limitations of the current approach. For example, our model is also not well suited for social graphs, where structure plays a minor role, and small-worldness implies that the subgraphs will span the majority of the input graph even at small radii. It should also be easy to modify our architecture to allow any number of graph kernels to be plugged in, allowing the network to learn which kernel is best suited for the dataset and task at hand. In principle, our architecture can also operate using graph kernels for both directed and undirected graphs (\eg GL and WL kernels), as well as edge attributed ones (\eg WL kernel, categorical attributes only). In general, node and edge features are not restricted to being categorical, however in the case of continuous features the kernel should be differential \wrt the features. \red{Note that these limitations in handling social graphs and the necessity for the kernel to be differential \wrt continuous features may limit the immediate applicability of our model in some domains.} Future work will aim to address these limitations and explore our network's performance on node classification and regression tasks.

\section*{Acknowledgments}
L.C. and A.B. are supported by the PRIN 2022 project n. 2022AL45R2 (EYE-FI.AI, CUP H53D2300350-0001). G.M. acknowledges financial support from the European Union - NextGenerationEU, in the framework of the iNEST - Interconnected Nord-Est Innovation Ecosystem (iNEST ECS$\_$00000043 – CUP H43C22000540006). In this regard, the views and opinions expressed are solely those of the authors and do not necessarily reflect those of the European Union, nor can the European Union be held responsible for them.



\bibliography{main}
\bibliographystyle{IEEEtran}

\begin{IEEEbiography}[{\includegraphics[width=1in,height=1.25in,clip,keepaspectratio]{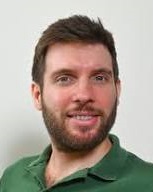}}]{Luca Cosmo} is an Assistant Professor in Computer Science at University Ca' Foscari of Venice, working in Geometric Deep Learning and Generative AI, with more than 50 published papers. He has been serving on the program committee and as a reviewer for top conferences and journals in these fields and he gave short courses and tutorials in geometry processing related topics at major conferences in the field. He is a member of ELLIS and co-founded a start-up working on geometry processing for industrial product inspection.

\end{IEEEbiography}

\begin{IEEEbiography}[{\includegraphics[width=1in,height=1.25in,clip,keepaspectratio]{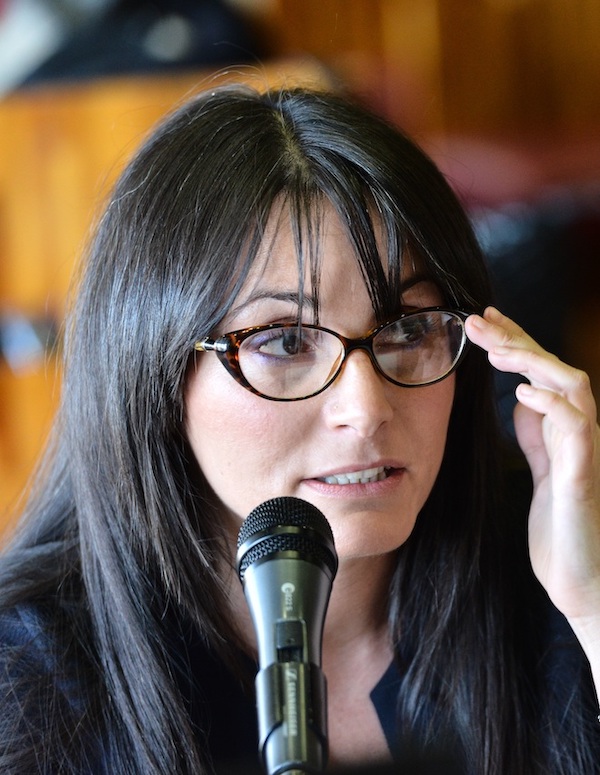}}]{Giorgia Minello}  Giorgia Minello worked in the industry for almost ten years before receiving, in 2019, her PhD in Computer Science from Ca’ Foscari, University of Venice (Italy), where she currently holds a Postdoctoral Research Fellow position. She is also a scientific collaborator for the IESE Business School, University of Navarra, in the context of a social network analysis project. Her research mainly focuses on Structural Pattern Recognition, Machine Learning and Natural Language Processing. The leading topic of her studies concerns the analysis of networks, such as the use of quantum approaches for structural analysis or the application of network analysis methods for mining textual corpora.
\end{IEEEbiography}

\begin{IEEEbiography}[{\includegraphics[width=1in,height=1.25in,clip,keepaspectratio]{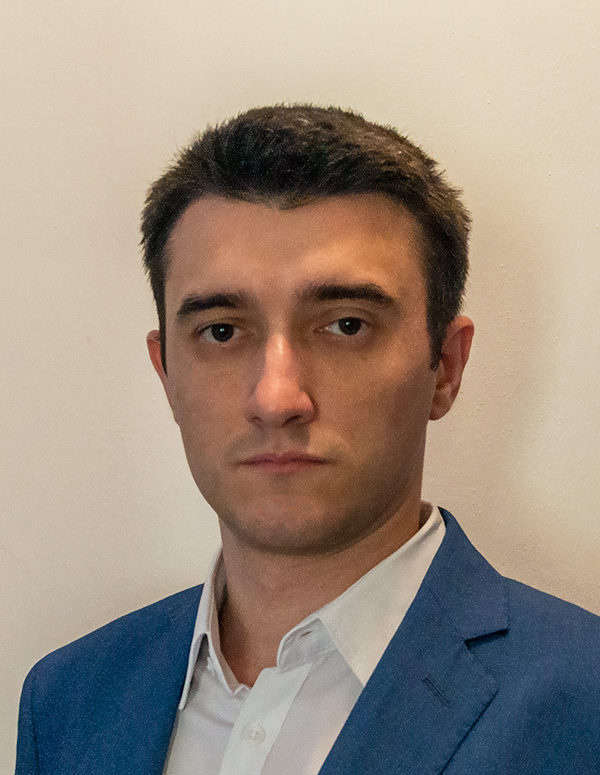}}]{Alessandro Bicciato}
received the PhD in computer science from the Ca' Foscari University of Venice, Italy in 2024. He is currently working as a post-doctorate researcher at the Computer Science Department of Ca' Foscari University of Venice, Italy. His research interests include network science and pattern recognition.
\end{IEEEbiography}

\begin{IEEEbiography}[{\adjincludegraphics[width=1in,height=1.25in,trim={{.12\width} 0 {.12\width} 0},clip,keepaspectratio]{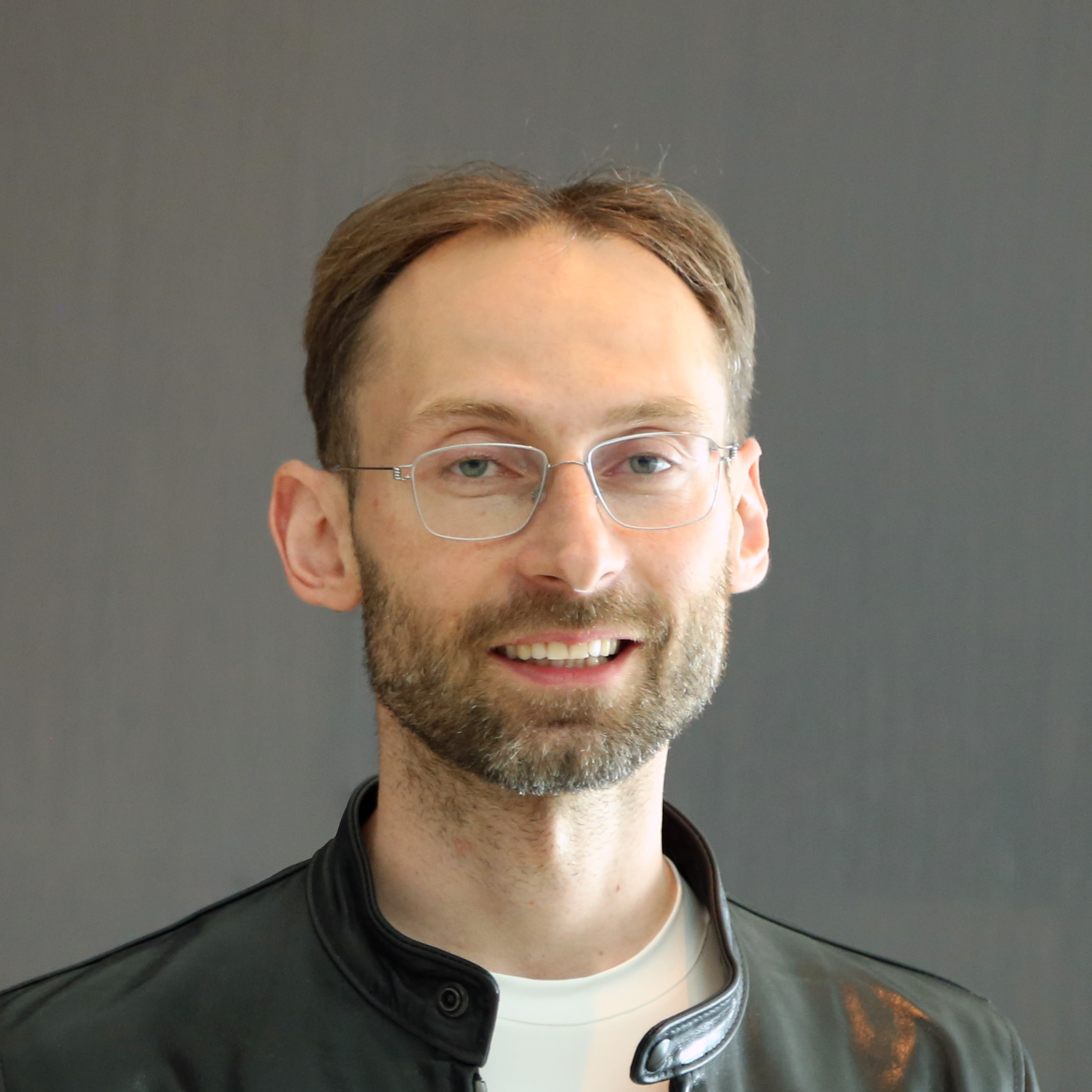}}]{Michael Bronstein}
Michael Bronstein is the DeepMind Professor of AI at the University of Oxford. He previously served as Head of Graph Learning Research at Twitter, professor at Imperial College London, and held visiting appointments at Stanford, MIT, and Harvard. He is the recipient of the Royal Society Wolfson Research Merit Award, Royal Academy of Engineering Silver Medal, Turing World-Leading AI Research Fellowship, five ERC grants, two Google Faculty Research Awards, and two Amazon AWS ML Research Awards. He is a Member of the Academia Europaea, Fellow of IEEE, IAPR, BCS, and ELLIS, ACM Distinguished Speaker, and World Economic Forum Young Scientist. In addition to his academic career, Michael is a serial entrepreneur and founder of multiple startup companies, including Novafora, Invision (acquired by Intel in 2012), Videocites, and Fabula AI (acquired by Twitter in 2019). He is the Chief Scientist at VantAI and scientific advisor at Recursion Pharmaceuticals. 
\end{IEEEbiography}

\begin{IEEEbiography}[{\adjincludegraphics[width=1in,height=1.25in,trim={{.12\width} 0 {.12\width} 0},clip,keepaspectratio]{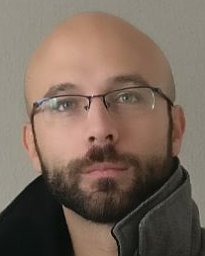}}]{Emanuele Rodol\`{a}}
is Professor of Computer Science at Sapienza University of Rome, where he leads the GLADIA group of Geometry, Learning and Applied AI, funded by an ERC Grant and a Google Research Award. He is currently an ELLIS fellow, and formerly fellow of the Humboldt Foundation and the Japan Society for the Promotion of Science. His research interests lie at the intersection of geometry processing, graph theory, geometric deep learning, applied AI and computer vision, and has published more than 100 papers in these areas.
\end{IEEEbiography}

\begin{IEEEbiography}[{\includegraphics[width=1in,height=1.25in,clip,keepaspectratio]{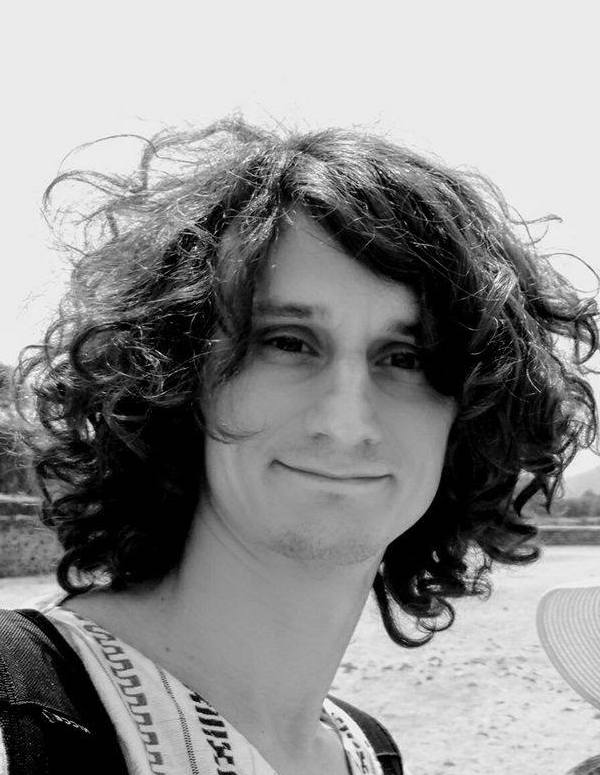}}]{Luca Rossi}
received the PhD degree in computer science from the Ca' Foscari University of Venice, Italy, in 2013. He is currently an Assistant Professor
with the Hong Kong Polytechnic University, having held various positions with the University of Birmingham, Aston University, Southern University of Science and Technology, and Queen Mary University of London. He has published more than 50 papers in international journals and conferences. His research interests include the areas of pattern recognition, data mining, and network science. He is currently a member of the editorial board of the journal Pattern Recognition and vice-chair of the technical committee 2 of the International Association for Pattern Recognition.
\end{IEEEbiography}

\begin{IEEEbiography}[{\includegraphics[width=1in,height=1.25in,clip,keepaspectratio]{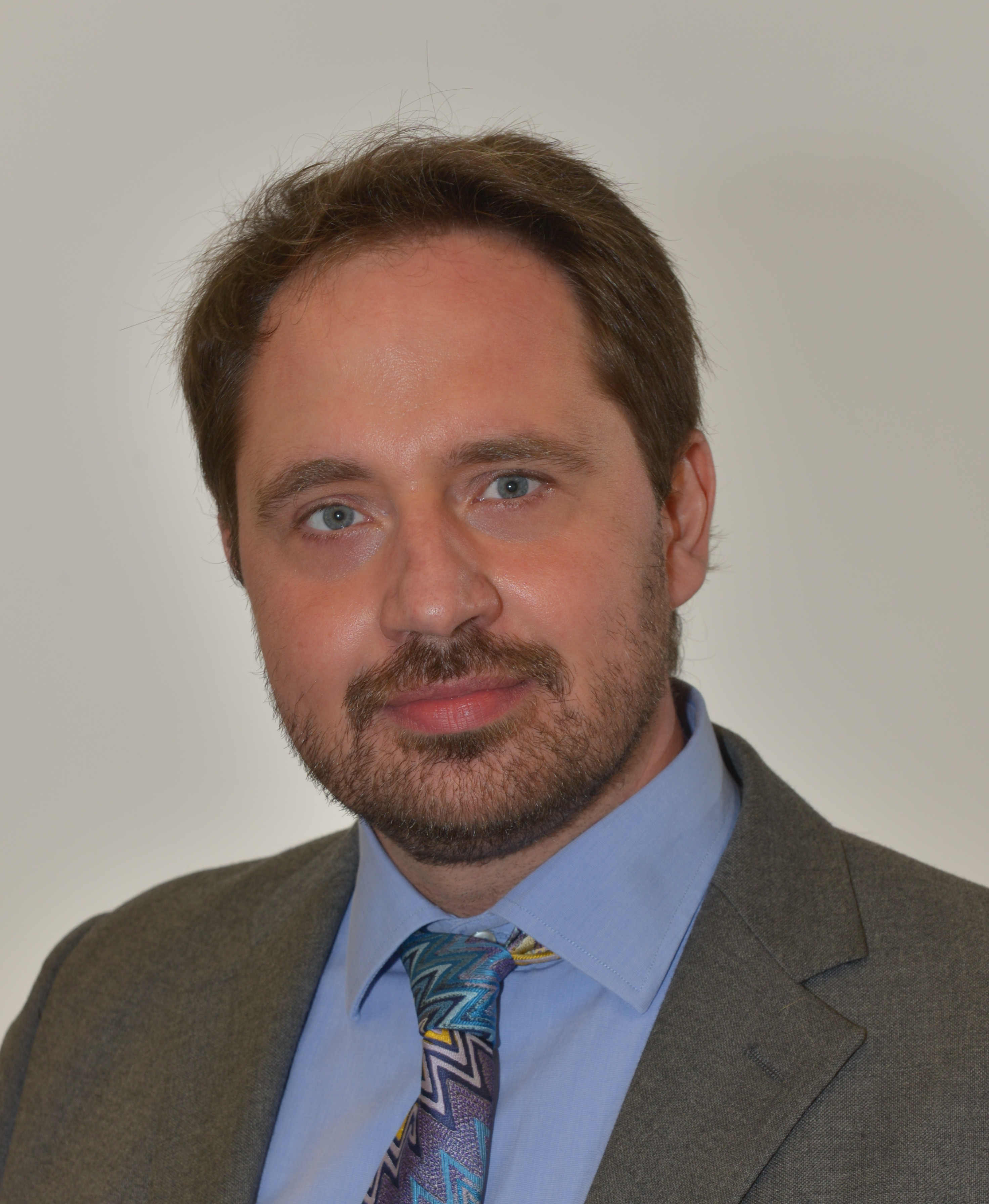}}]{Andrea Torsello}
received his PhD in computer science at the University of York, UK. From 2007 he is with Ca’Foscari University of Venice, Italy, where he is Full Professor. 
He has published over 150 papers in international journals and conferences. His research interests are in the areas of Computer Vision and Pattern Recognition, in particular the interplay between Stochastic and Structural approaches as well as Game-Theoretic and Physical models. He is currently a member of the editorial board of the journal Pattern Recognition and chair of the technical committee 2 of the International Association for Pattern Recognition.
\end{IEEEbiography}

\end{document}